\documentclass[journal, twocolumn]{IEEEtran}
\usepackage[pdftex]{graphicx}

\usepackage{amsmath}
\usepackage{array,booktabs}
\usepackage{algorithmic}
\usepackage{url}
\usepackage{cite}
\usepackage{multirow}
\usepackage{balance}
\usepackage{xcolor}
\usepackage{subcaption}

\usepackage{balance}

\begin{document}

\title{Automated Pest Detection with DNN on the Edge for Precision Agriculture}
	

\author{
	Andrea~Albanese, 
	Matteo~Nardello,~\IEEEmembership{Member,~IEEE,} and Davide~Brunelli,~\IEEEmembership{Senior Member,~IEEE}\\
\thanks{This work was supported by the Italian Ministry for Education, University and Research (MIUR) under the program “Dipartimenti di Eccellenza (2018-2022)”. Moreover, this research was supported by TIM Telecom Italia S.p.A.}
\thanks{A. Albanese, M. Nardello, D. Brunelli are with the Department of Industrial Engineering DII, University of Trento, 38123 Trento, Italy (e-mail: \{name.surname\}@unitn.it).}
\thanks{Citation information: DOI 10.1109/JETCAS.2021.3101740, IEEE Journal on Emerging and Selected Topics in Circuits and Systems. This article has been accepted for publication in a future issue of this journal, but has not been fully edited. Content may change prior to final publication.}
\thanks{Personal use is permitted, but republication/redistribution requires IEEE permission. See \url{http://www.ieee.org/publications_standards/publications/rights/index.html} for more information.}
}

\maketitle

\begin{abstract}
Artificial intelligence has smoothly penetrated several economic activities, especially monitoring and control applications, including the agriculture sector. However, research efforts toward low-power sensing devices with fully functional machine learning (ML) on-board are still fragmented and limited in smart farming. Biotic stress is one of the primary causes of crop yield reduction. With the development of deep learning in computer vision technology, autonomous detection of pest infestation through images has become an important research direction for timely crop disease diagnosis. This paper presents an embedded system enhanced with ML functionalities, ensuring continuous detection of pest infestation inside fruit orchards. The embedded solution is based on a low-power embedded sensing system along with a Neural Accelerator able to capture and process images inside common pheromone-based traps. Three different ML algorithms have been trained and deployed, highlighting the capabilities of the platform. Moreover, the proposed approach guarantees an extended battery life thanks to the integration of energy harvesting functionalities. Results show how it is possible to automate the task of pest infestation for unlimited time without the farmer's intervention.
\end{abstract}


\section{Introduction}
Due to the constant growth of food production demand, in Europe, agriculture is responsible for more than 10\% of greenhouse gas emissions and 44\% of water consumption nowadays. Chemical treatments (e.g., pesticides) are being intensively used to improve the market penetration of fruit crops, leading to a remarkable impact on pollinators and the planet's ecosystem. Thus, there is an increasing interest in new techniques to lower the water demand~\cite{sartori16} and optimize pesticide treatments to preserve natural resources\footnote{Source: https://horizon-magazine.eu/article/how-crop-and-animal-sensors-are-making-farming-smarter.html}.

Farmers and researchers have been teaming up to develop smart systems for precision agriculture. Networks of smart sensors are mounted directly inside fruit and vegetable orchards, and advanced machine learning algorithms optimize the agriculture resources usage, enabling crops monitoring by gathering real-time data about the orchard's health.

Usually, pheromone-based traps are equipped with a passive camera that captures pictures of the trapped insects and sends them through internet nodes to the cloud. Afterward, an expert, or a farmer, is required to review the captured images to check the effective presence of dangerous parasites and eventually plan a local counteraction (i.e., a pesticide treatment). However, this process requires a high amount of data to be sent over long-range communication, making the application inefficient and time-expensive due to regular human intervention in the detection process. 

Recently, researchers have started investigating smart-trap usage for pest detection as a solution to increase the wealth of orchards while lowering pesticide demand~\cite{brunelli2019energy}. These traps -- installed directly inside orchards -- can autonomously detect dangerous parasites and alert the farmer to apply targeted pesticide treatments. Thanks to the implementation of sophisticated at-the-edge machine learning algorithms, the smart trap can detect dangerous parasites without remote cloud infrastructure as generally required for machine learning applications. 

This solution opens many new possibilities for monitoring application in precision agriculture: i) The optimized usage of the limited energy available inside orchards; ii) The balanced distribution of the whole implemented application by exploiting the computation capabilities at different levels (edge-concentrators-cloud) for achieving better scalability; iii) The capability of using the recent low-power long-range and low-data-rate radio protocols~\cite{Polonelli19slotted}, as we do not need to send the acquired massive image data but only the result of its analysis; iv) Finally, thanks to a low energy budget, the capabilities of exploiting energy harvesting extending to unlimited lifetime the smart traps.

This paper presents a smart trap for pest detection running a Deep Neural Network on edge. The smart trap enables fast detection of pests in apple orchards by using ML algorithms that improve the overall system efficiency~\cite{mudassar2019camel,shoaran2018energy}. All the computation is done on-the-node, thus slimming down the amount of data transmission and limiting it to a simple notification of few bytes if threats are detected. The smart trap features an energy harvesting system composed of a real-time clock (RTC) to trigger the pest detection task -- implemented using a low-power STM32L0\footnote{https://www.st.com/en/microcontrollers-microprocessors/stm32l0-series.html} MCU --, a small Li-Ion battery, and a solar panel to power its operations indefinitely. 
The hardware solution is developed on top of a RaspberryPi microcomputer as a mainboard equipped with a camera as an image sensor and an Intel Neural Compute Stick (NCS) as a neural-accelerator to optimize the inference execution and, accordingly, the energy consumption~\cite{motaman2019perspective,gao2020edgedrnn,dey2019pre}.
The overall system has been characterized by considering its power consumption over a full application cycle to find out the power-hungry tasks that require dedicated power optimization to meet the system energy balance. Furthermore, three different CNN models for image classification have been compared: a modified LeNet-5~\cite{guo2017simple} which, thanks to its straightforward architecture, can speed the execution up, a VGG16~\cite{simonyan2014very, qassim2018compressed} model which features a more deep CNN architecture capable of achieving better recognition accuracy when classifying class of objects, and finally a MobileNetV2~\cite{Sandler_2018_CVPR} network perfectly suitable to be evaluated on the edge by resource constrained platforms.

This paper makes three main contributions: 
\begin{itemize}
    \item 
        The development and characterization of an IoT smart traps that can be deployed and left working unattended for prolonged times without the need for any human intervention;
    \item
        The study, training, optimization and validation of three different ML models suitable for the resource-constrained embedded platform. Results and performance comparison are presented and determine the model used for the final deployment.  
    \item
        The platform sustainability assessment when powered using the solar energy harvester.
\end{itemize}

The rest of the paper is organized as follows. In Section~\ref{relate} related works are discussed. Section~\ref{DNN} present a review of the three machine learning algorithms investigated. The overall system architecture is presented in Section~\ref{arch}, and the main design choices are discussed. Section~\ref{res} presents the system's performance and discusses the achieved results. Finally, Section~\ref{close} presents the final remarks.

\section{Related work}
\label{relate}
The recent advances in smart agriculture are mainly powered by the progress in wireless sensor networks (WSNs), unmanned autonomous vehicles (UAVs)~\cite{Tosato19swarm}, machine learning, low-power imaging systems~\cite{Nardello19camaroptera}, and cloud computing. Tiny sensors are deployed in difficult-to-access areas for the periodic acquisition of in-field data~\cite{Polonelli18euc}. Although research and comparable prototypes are still limited, we provide a discussion on successful automating pest detection tasks.
              
Monitoring is a crucial component in pheromone-based pest control systems~\cite{carde1995control, Witzgall2010}. In widely used trap-based pest monitoring, captured digital images are analyzed by human experts for recognizing and counting pests. Manual counting is labor-intensive, slow, expensive, and error-prone, which precludes real-time performance and cost targets.

Deep learning techniques are recently used to overcome these limitations and achieve a completely automated, real-time pest monitoring system by removing the human from the loop~\cite{segalla2020neural}. \cite{agriculture10050161} is one of the recent works that exploit ML techniques to classify insects. These works can be grouped based on the applied method. In terms of image sources, many articles have investigated insect specimens ~\cite{KANG2012431, KANG2014143, WEN2009299}. Unfortunately, even if specimens are usually well preserved and managed in a laboratory environment, this approach is not suitable for creating a model for image classification when data is collected in the wild. Researchers have thus proposed to extend the datasets with images captured from real traps ~\cite{liu2019pestnet, agriculture10050170, agriculture10050161}. 

\begin{figure}[b]
	\centering
	\includegraphics[width=0.75\columnwidth]{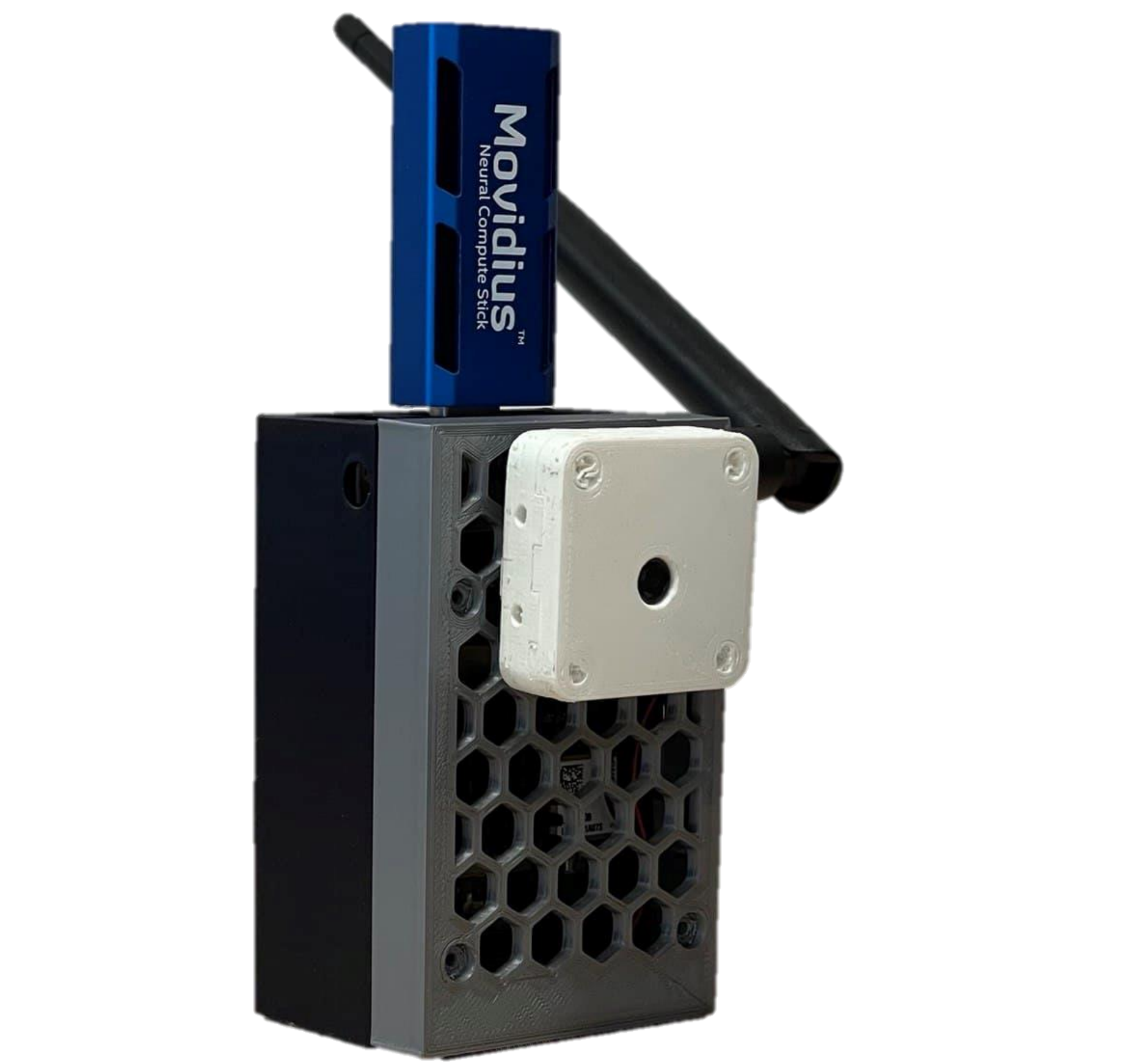}
	\caption{An example of the assembled prototype used during indoor testing.}
	\label{fig:prot}
\end{figure}

From an algorithmic perspective, various types of features have been used for insect classification, including wing structures~\cite{KANG2012431, KANG2014143, WEN2009299}, morphometric measurement~\cite{WANG2012102} and global image features~\cite{LIU201682 ,YAO2012978, DING201617, XIE2018233}.

Different classifiers were also developed starting from the various feature extraction methods, including but not limited to support vector machines (SVM) \cite{WANG2012102, 7873750, OMRANI2014512}, artificial neural networks (ANN) \cite{Kaya2014, doi:10.1080/03235408.2013.763620, samanta2012tea, ESPINOZA2016495} and k-nearest neighbors (KNN) \cite{qiao2018automated, COMO2017438}. In general, however, these proposed methods were not tested under real application scenarios, for example, to classify real-time images acquired from real traps deployed inside orchards.

To solve some of these early methods' problems, more recently, Deep Learning has been proposed in the literature~\cite{plants9101302, agriengineering2030029, 10.1145/3232651.3232661}. Their variants have emerged as the most effective method for object recognition and detection by achieving state-of-the-art performance on many well-recognized datasets also in the domain of precision agriculture. A particular class of DL algorithms -- well known as Convolutional Neural Network (CNN) -- has made a clear breakthrough on computer vision techniques for pest detection. Many sorts of algorithms based on CNN have emerged, significantly improving current systems' performance for classification as well as object localization~\cite{ai1020013, Patel_2021, 10.1145/3209914.3209945, 7112511, Uijlings2013}.  

Inspired by this research line, we adopt a Region-based detection pipeline with convolutional neural networks as the image classifier to classify in-situ images captured inside pheromone-based traps deployed inside apple orchards. The raw images are firstly preprocessed with color correction. 

Then, trained ConvNets are applied to densely sampled image patches to extract single regions of interest (ROIs). ROIs are then filtered by non-maximum suppression, after which only those with probabilities higher than their neighbors are preserved. Finally, the remaining ROIs are thresholded. ROIs that meet the threshold are considered as proposed detections. 

Even if the approach presented in this paper is not a novelty, the proposed solution can provide state-of-the-art detection results directly from apple orchards. Our system shows a suitable accuracy for the implemented task without first uploading acquired data to the cloud. All the computation is carried out autonomously on the edge, with the capability to exploit energy harvesting to avoid the energy overhead of the metering infrastructure and extend the lifetime of the installation.

\section{System architecture}
\label{arch}
The system has been designed to bring IoT technologies into a domain that requires data collection over vast areas. In this scenario, long-range communication, and high scalability from multiple devices. Thanks to the onboard recognition algorithm, the smart trap's data transmission is limited to a few bytes, thus exploitable in any rural area.  Only the result of the inference will be transmitted, making it suitable even for low bitrate communication. If the farmer needs a visual confirmation from the captured picture, a few images per day can be transmitted. Figure~\ref{fig:prot} presents the system prototype.

\subsection{Hardware Implementation}
Each smart trap is built on a custom hardware platform that includes: a small, low-power image sensor to collect images; a compact single board computer from Raspberry PI; an Intel Neural Compute module for optimizing the execution of the ML task; a long-range radio chip for communication; and a solar energy-harvesting power system for collecting and storing energy from the environment.  Figure~\ref{fig:harvester} presents the schematic block overview of the proposed IoT device. Its main functionalities are designed as follows. 

\subsubsection{\textbf{Sensing}}
The smart trap uses a Sony IMX219 image sensor. The IMX219 is a low-power Back-illuminated CMOS image sensor. 
The sensor utilizes 1.12 µm pixel technology that offers high sensitivity and only needs a few external passive components to work. It integrates black-levels calibration circuit, automatic exposure, and gain control loop to reduce host computation and commands to the sensor to optimize the system power consumption.

\subsubsection{\textbf{Processing}}
The processing layer is mainly composed of two parts. The first is a Raspberry minicomputer to manage the sensor acquisition, processing the captured pictures, and the transmission. After several tests with different releases on the market, we select the Pi3 version as the best trade-off between computing capability, energy demand, and cost. The second part consists of a neural accelerator, namely an Intel Neural Compute Stick, that is activated only during the ML task and reduces the inference time.

\subsubsection{\textbf{Transmission}}
The smart trap is equipped with a LoRa module~\cite{Tessaro18lora}. The connectivity is provided by a RFM95W transceiver featuring a LoRa low-power modem and a +20 dBm power amplifier that can achieve a sensitivity of over -148 dBm. The LoRa IC is then connected to an external antenna with a maximum gain of 2 dBi. 

\subsubsection{\textbf{Power unit}}
\begin{figure}[t]
	\centering
	\includegraphics[width=0.95\columnwidth]{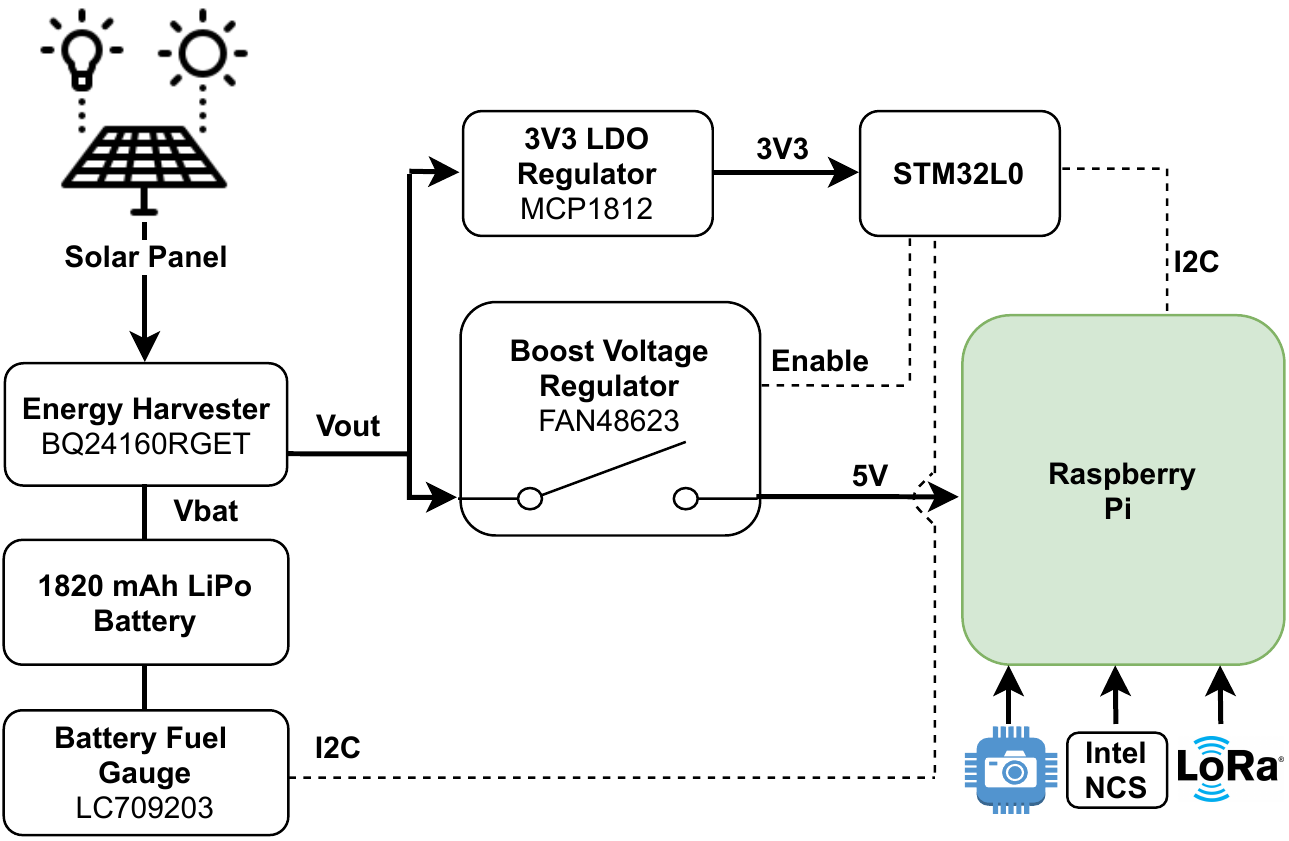}
	\caption{Solar energy harvester and power management circuit schematic block.}
	\label{fig:harvester}
\end{figure}

The smart trap integrates a complete power supply with energy harvesting functionalities to efficiently use the LiPo battery's limited energy. Figure~\ref{fig:harvester} presents the schematic block of the power supply module. Starting from the top left corner, the solar panel is connected directly to the energy harvester BQ24160, used to recharge a 1820mAh Li-Po battery. Two voltage converters are connected to it: the first, a  MCP1812\footnote{https://www.microchip.com/wwwproducts/en/MCP1812} LDO, is in charge to generate 3.3V to the external microcontroller. The second, a FAN48623\footnote{https://www.onsemi.com/pdf/datasheet/fan48623-d.pdf} Boost converter, provides a stabilized 5V to the Raspberry. This converter is controlled by an STM32L0 MCU and enabled according to the implemented power policy (e.g., SW programmed intervals).  A battery fuel gauge LC709203F\footnote{https://www.onsemi.com/products/power-management/battery-management/battery-fuel-gauges/lc709203f} is used by the MCU for battery status monitoring. The external low-power STM32L0 MCU is in charge of the energy management of the platform. It enables the power-up and shutdown of the Raspberry and manages the harvesting functionalities, guaranteeing the optimal battery life without the farmer's intervention.

\subsection{Detection pipeline}
\begin{figure}[t]
	\centering
	\includegraphics[width=0.9\columnwidth]{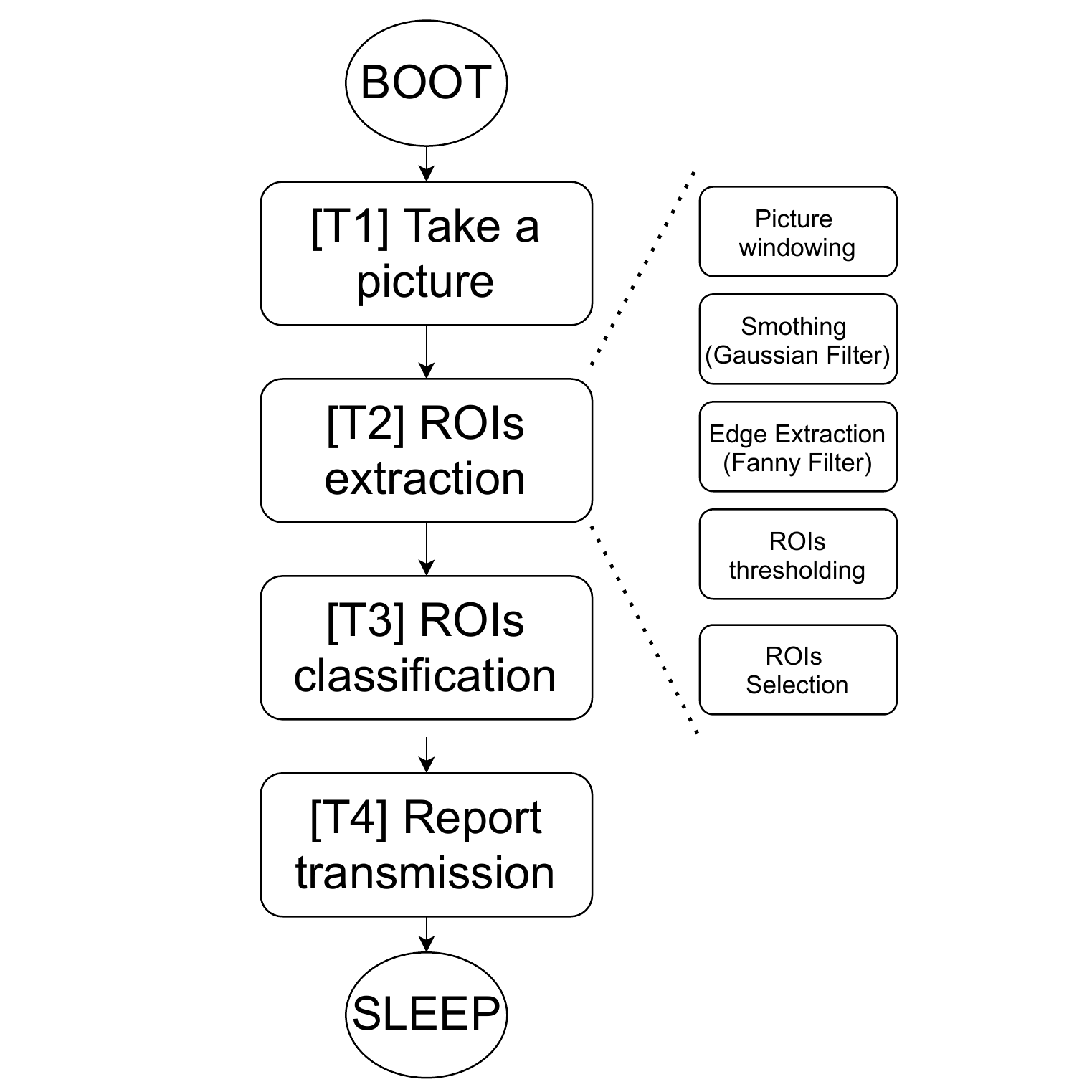}
	\caption{Detection Pipeline work flow. On the right, the RoIs extraction procedure is highlighted.}
	\label{fig:flow}
\end{figure}

A pipeline is a set of processing elements that describes the data flow and defines how the information is processed along the way.  In our case, the smart trap implements a multi-stage pipeline that processes an image after its capture to extract dangerous insects, called Codling Moth. The automatic detection pipeline is shown in Figure~\ref{fig:flow}. 

We use sliding windows and a trained image classifier. The classifier is applied to local windows at different locations of the entire image to extract and classify the regions of interest (ROIs). ROIs are a portion of the captured image encompassing just a single insect. An example of this operation is depicted in Figure~\ref{fig:dataset}.

The classifier’s output is a single scalar, representing the probability that a particular ROI contains a codling moth. 

These ROIs are regularly and densely arranged over the image and thus largely overlapping. Therefore, we perform Smoothing (or blurring) of the frame with a Gaussian filter and then Edge extraction through the Canny operator to select only the ROIs whose respective probability is locally maximal. 

After this detection phase, ROIs are further analyzed by the learning algorithm that tries to assert if the detected insect is a codling moth or not. The final output of the detection pipeline is the initial captured images along with the colored square highlighting the detected positive ROIs, as shown in Figure~\ref{fig:output}.

After the DNN assessment, a report is generated and sent using the LoRa modem to the farmer. 

\subsection{Edge Accelerator}

    Edge accelerator is a class of purpose-built System on a Chip (SoC) for running deep learning models efficiently on edge devices. Different companies have proposed hardware solutions to accelerate the execution of deep learning algorithms at the edge of the network. To this end, we took a systematic look at a set of edge accelerators, their working principles, and performance in terms of executing different learning tasks. We compared three different platforms: Intel NCS2, Google Coral USB TPU and Nvidia Jetson Nano, which are state of the art on the market. 
    \\
    \textbf{Energy consumption.} Energy is a precious resource in battery-powered edge accelerators. From an energy consumption viewpoint, the most power-hungry platform is the Jetson Nano, requiring up to 10W when exploiting the GPU during the inference\footnote{https://docs.nvidia.com/jetson/l4t/index.html\#page/Tegra\%20Linux\%20Driver\\\%20Package\%20Development\%20Guide/power\_management\_nano.html}. On the contrary, the power consumption of the Google TPU is around 5W. A similar power need is measured for  the Intel NCS 2 along with the carrier board (in our case, a RPi 3B+)~\cite{libutti2020benchmarking}, mainly divided into 2W consumed by the accelerator and 3W by the RPi. \\
    \textbf{Performance.} The execution time of the model inference is a key metric for sensory systems. Among the 3 platforms compared, the Nvidia Jetson presents the higher computational capabilities, followed by the Google TPU~\cite{10.1145/3363347.3363363}. The Intel NCS 2 is the less powerful platform, but still perfectly suitable for the proposed implementation that does not require hard-real-time execution of the ML task. In our case, we are more focused on energy reduction; thus we selected the Intel NCS2 as a neural accelerator for the proposed application. \\
    \textbf{Compatibility.} Although Edge TPU appears the most competitive in terms of performance and size, it is also the most limiting in software as only TensorFlow frameworks are supported. Moreover, it does not support the full TensorFlow Lite but only the models that are quantized to 8-bits integer (INT8). This contrasts with NCS2 that also supports FP16/32 (16/32-bit floating point) in addition to INT8. In addition, the Intel NCS2 is widely supported by the community, thanks to the OpenVINO\footnote{https://software.intel.com/content/www/us/en/develop/tools/openvino-toolkit.html} toolkit that allows the conversion of different Machine learning frameworks. Nvidia’s software is the most versatile as its TensorRT supports most ML frameworks including MATLAB. Google TPU and NCS2 are designed to support some subset of computational layers (primarily for computer vision tasks), but Jetson Nano is essentially a GPU, and it can do the similar computations as its big brother desktop GPUs.\\
    \textbf{Availability.} Hardware availability was also a factor limiting the platforms and configuration tested. At the time of our project kickoff, only the Intel NCS2 and the Google TPU were available and adequately documented.  
    \\
    The exploration of the design space, done by evaluating the three neural accelerators, suggested continuing our development on the Intel NCS2, because it represents the best trade-off in terms of performance, energy consumption and learning model’s compatibility.  

\section{Deep neural networks}
\label{DNN}
\begin{figure}
		\centering
		\includegraphics[width=0.9\columnwidth]{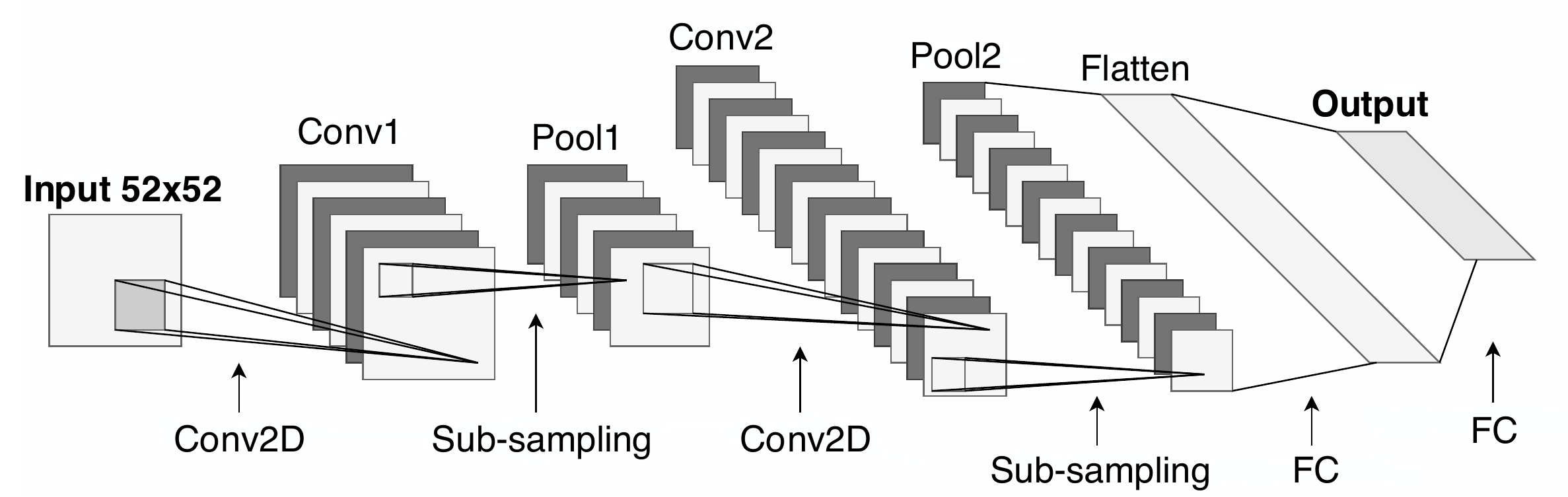}
		\caption{LeNet-5 architecture.}
		\label{fig:lenet}
\end{figure}

\begin{figure}
		\centering
		\includegraphics[width=0.9\columnwidth]{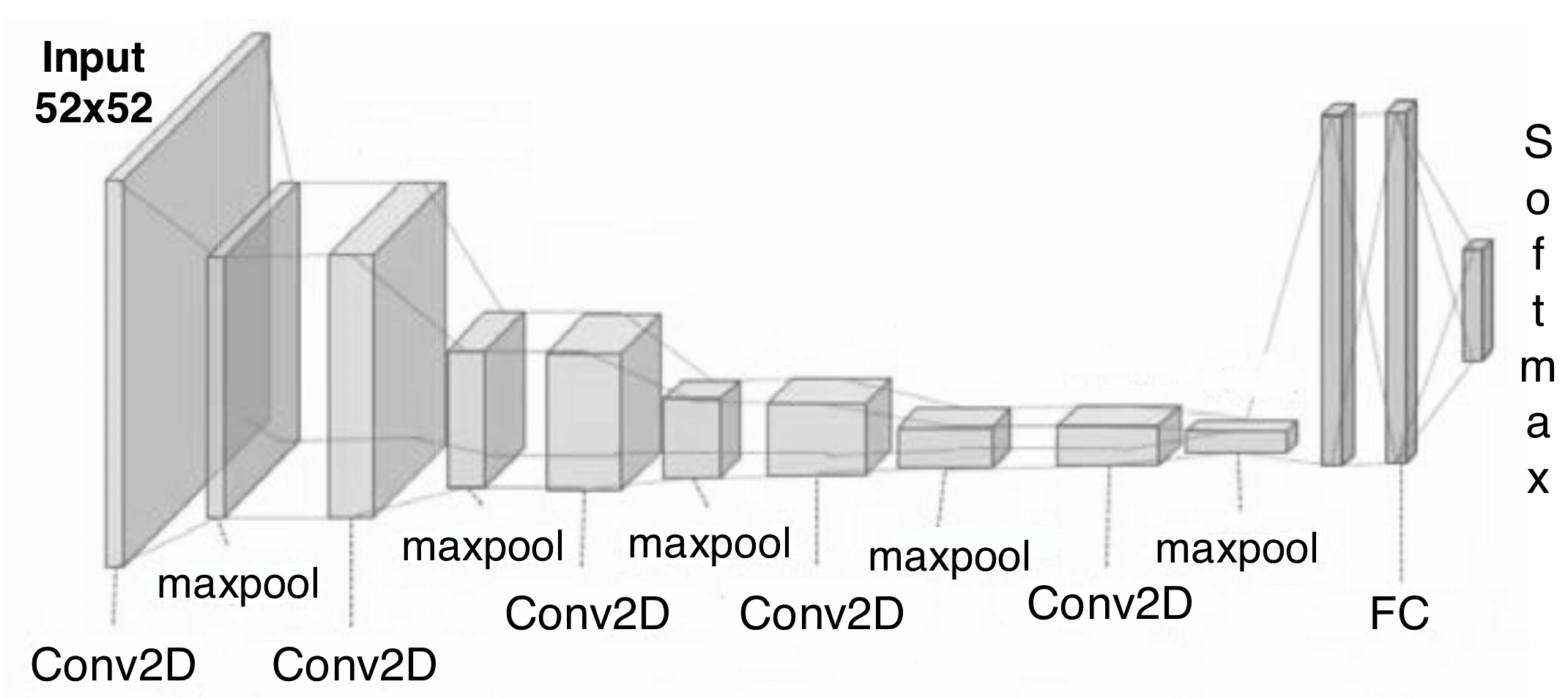}
		\caption{VGG16 architecture.}
		\label{fig:vgg}
\end{figure}

\begin{figure}
		\centering
		\includegraphics[width=0.8\columnwidth]{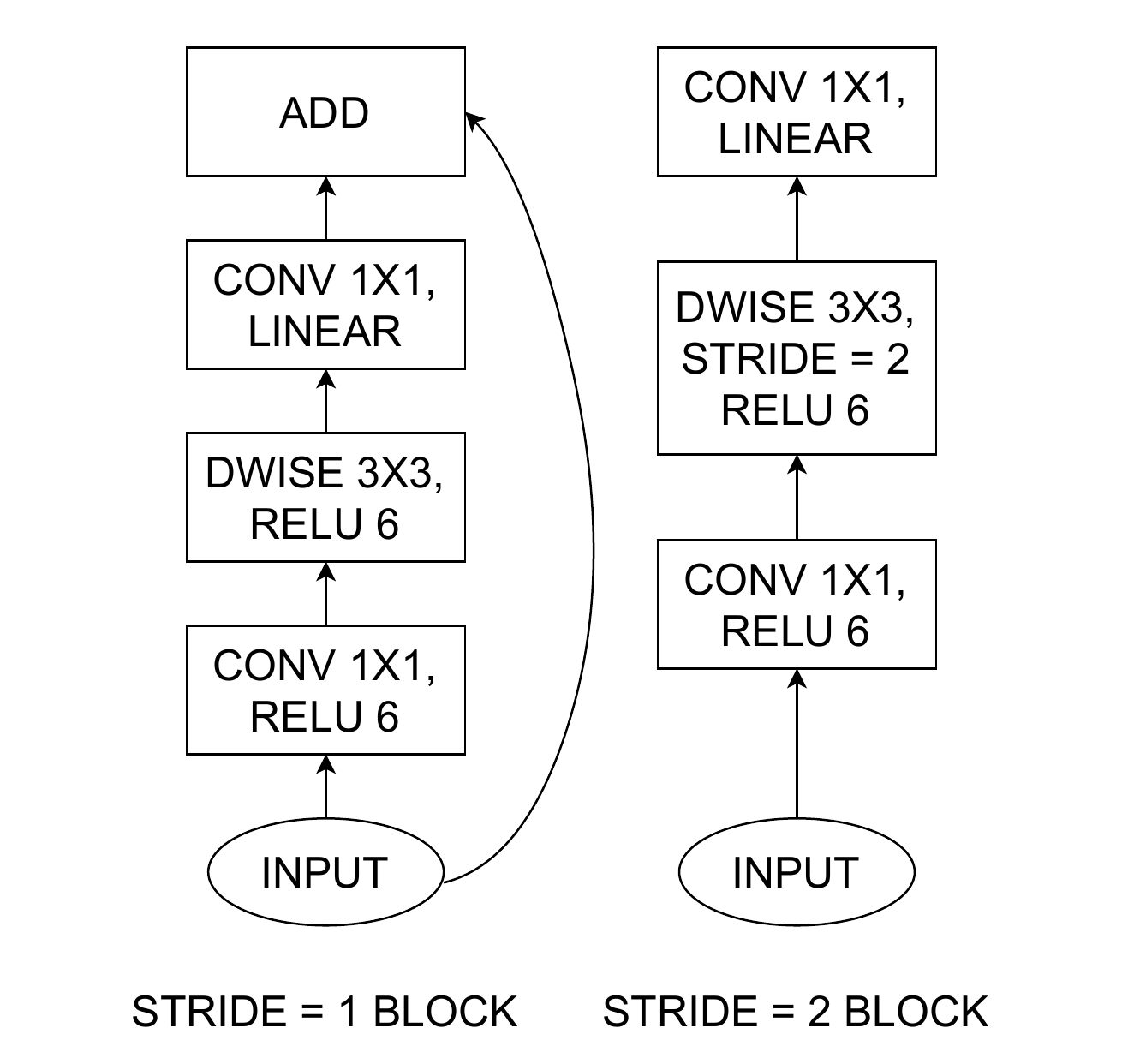}
		\caption{MobileNetV2 architecture.}
		\label{fig:MN2}
\end{figure}

Deep Learning is a class of machine learning algorithms based on the so called ANNs trained through feature learning techniques. DL algorithms can improve the recognition capability of many systems by simulating the biological neural networks (i.e., the human brain behaviour). They can automatically learn features at multiple levels of abstraction and compose them to learn complex ones.
For this application purpose, three state-of-the-art DNN architectures represented in Figures~\ref{fig:lenet},~\ref{fig:vgg} and~\ref{fig:MN2} has been chosen:
\begin{itemize}
    \item 
        A modified \textbf{LeNet-5}~\cite{guo2017simple}, presented in Figure~\ref{fig:lenet},  which features a simple and straightforward structure. It has been designed for hand-written character recognition, but we extended for classification problems with few modifications. As revealed in Figure~\ref{fig:lenet}, it is composed of seven layers: 3 convolutional, 2 subsampling and one fully connected layer followed by a softmax classifier. Moreover, the second convolutional block does not use all the features produced by the average pooling layer. This permits the convolutional kernels to learn different patterns and improve the classification accuracy. It also makes the network less computing demanding, which turns suitable for embedded platforms. By changing the original activation function (i.e., \textit{tanh}) to a rectified linear unit one, it is possible to extend this network for classification tasks where specific patterns have to be recognized (especially for our case of pest detection).
    \item 
        \textbf{VGG16}~\cite{simonyan2014very, qassim2018compressed}, represented in Figure~\ref{fig:vgg},  is characterized by a complex architecture and, when compared with LeNet performance, reveals the higher potential in classification accuracy. It uses convolutional kernels 3$\times$3 with stride 1 and the same padding and max-pool layers of 2$\times$2 filter and stride 2. In this way, convolutional kernels can learn different patterns and geometrical shapes. Fully connected layers enable the classification of objects depending on the position of shapes in the image. Thus, it makes this network perfect for recognition problems as for this application purpose.
    \item 
        \textbf{MobileNetV2}~\cite{Sandler_2018_CVPR}, presented in Figure~\ref{fig:MN2}. It is based on an inverted residual structure where the shortcut connections are between the thin bottleneck layers. In MobileNetV2, there are two types of inverted residual blocks. One is with stride of 1 and one with stride of 2 for downsizing. The network is composed by 3 layers for both types of blocks. The first layer is 1×1 convolution with ReLU6. The second layer is the depthwise convolution. The third layer is another 1×1 convolution but without any non-linearity. The intuition is that the bottlenecks encode the model’s intermediate inputs and outputs while the inner layer encapsulates the model’s ability to transform from lower-level concepts such as pixels to higher-level descriptors such as image categories. Thanks to this architecture, the network is perfectly suitable to build highly efficient mobile models. Finally, as with traditional residual connections, shortcuts enable faster training and better accuracy.
        
\end{itemize}

We tested these three learning model architectures to select the best trade-off between complexity, accuracy, and power consumption. 

\subsection{Training session}
The dataset for the training phase has been constructed in a semi-automatic way thanks to a specific image processing algorithm. It starts from raw pictures, extracts relevant features such as contours and blobs, and generates tiles with codling moth or general insects for the dataset. Initially, the overall dataset consists of 1100 images where 30\% are used for validation. 

\begin{figure}[]
  \centering
  \begin{subfigure}[b]{0.85\linewidth}
    \includegraphics[width=\linewidth]{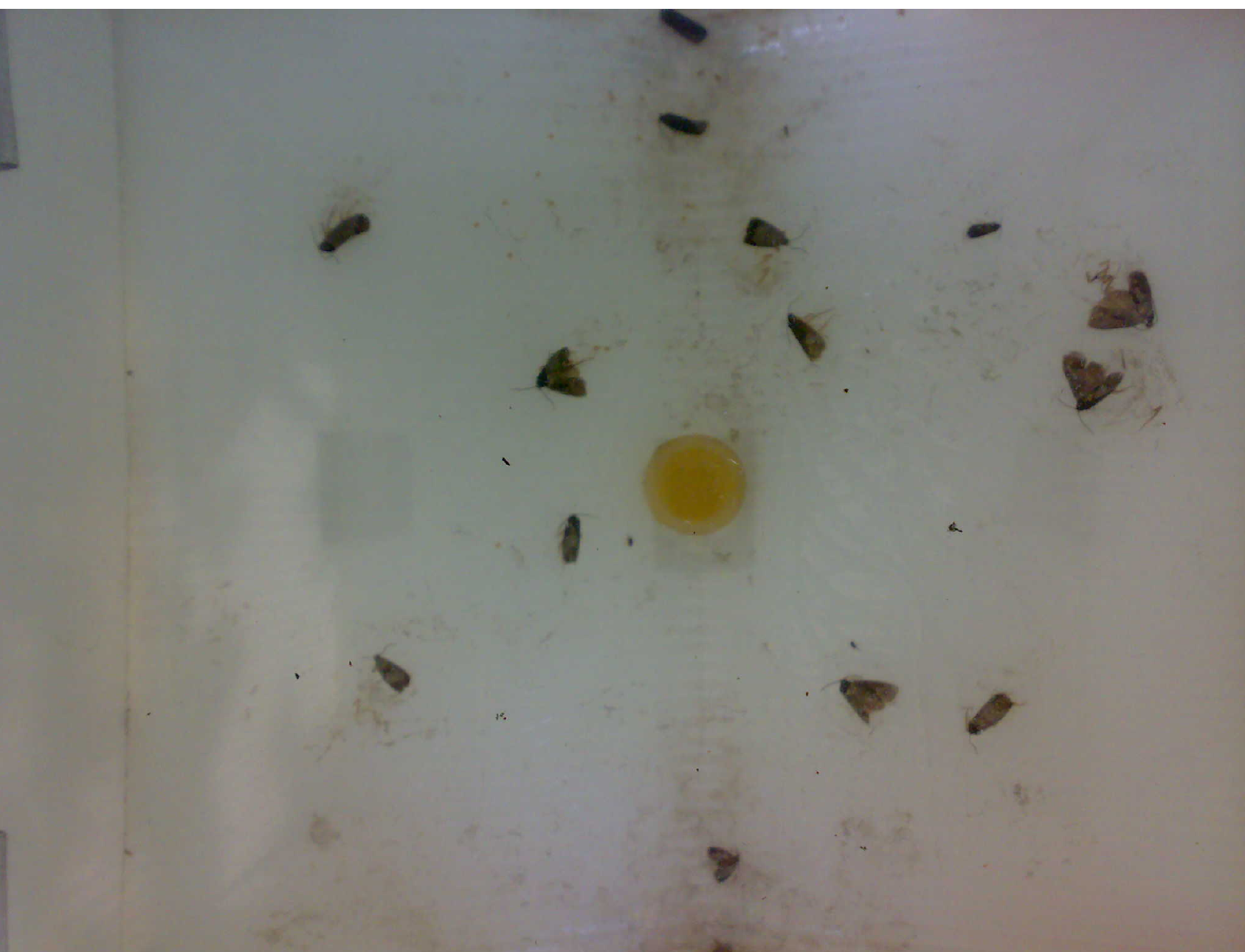}
     \caption{Raw picture.}
     \label{raw}
  \end{subfigure}
  \begin{subfigure}[b]{0.42\linewidth}
    \includegraphics[width=\linewidth]{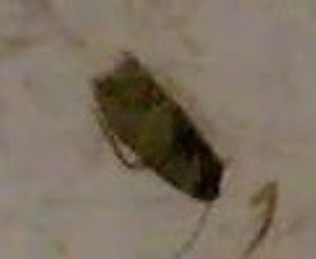}
    \caption{RoI with codling moth.}
    \label{codling}
  \end{subfigure}
  \begin{subfigure}[b]{0.42\linewidth}
    \includegraphics[width=\linewidth]{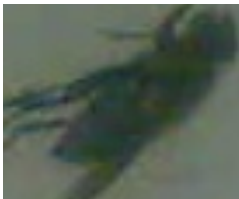}
    \caption{RoI with general insect.}
    \label{insect}
  \end{subfigure}
  \caption{Image (a) presents a photo taken inside a pheromone based trap by the proposed smart camera while (b) and (c) present an example of extracted ROIs with a single insect during the pre-processing phase.}
  \label{fig:dataset}
\end{figure}

The number of dataset images has been increased by combining augmentation techniques such as dataset generation (before training) and in-place augmentation (during training). These techniques artificially expand the dimension of a training dataset and improve the dataset sparsity to prone the network to generalization capability. These techniques' purpose is to create an expanded version of the original dataset by applying various image processing operators (e.g., shift, flips, zooms). Finally, the overall dataset consists in 4400 images divided into two classes: 3200 for \textit{codling moth} and 1200 for \textit{general insect}. It has been further split into 3500 images for train (2500 for \textit{codling moth} and 1000 for \textit{general insect})  and 900 for test (625 for \textit{codling moth} and 275 for \textit{general insect}). An example of the images is shown in Figure~\ref{fig:dataset}, depicting a raw picture used for the dataset construction (Figure~\ref{raw}), a tile with Codling Moth (Figure~\ref{codling}), and a tile with a general insect (Figure~\ref{insect}) used for the network training.

\begin{figure*}[]
  \centering
  \begin{subfigure}[b]{0.32\linewidth}
    \includegraphics[width=\linewidth]{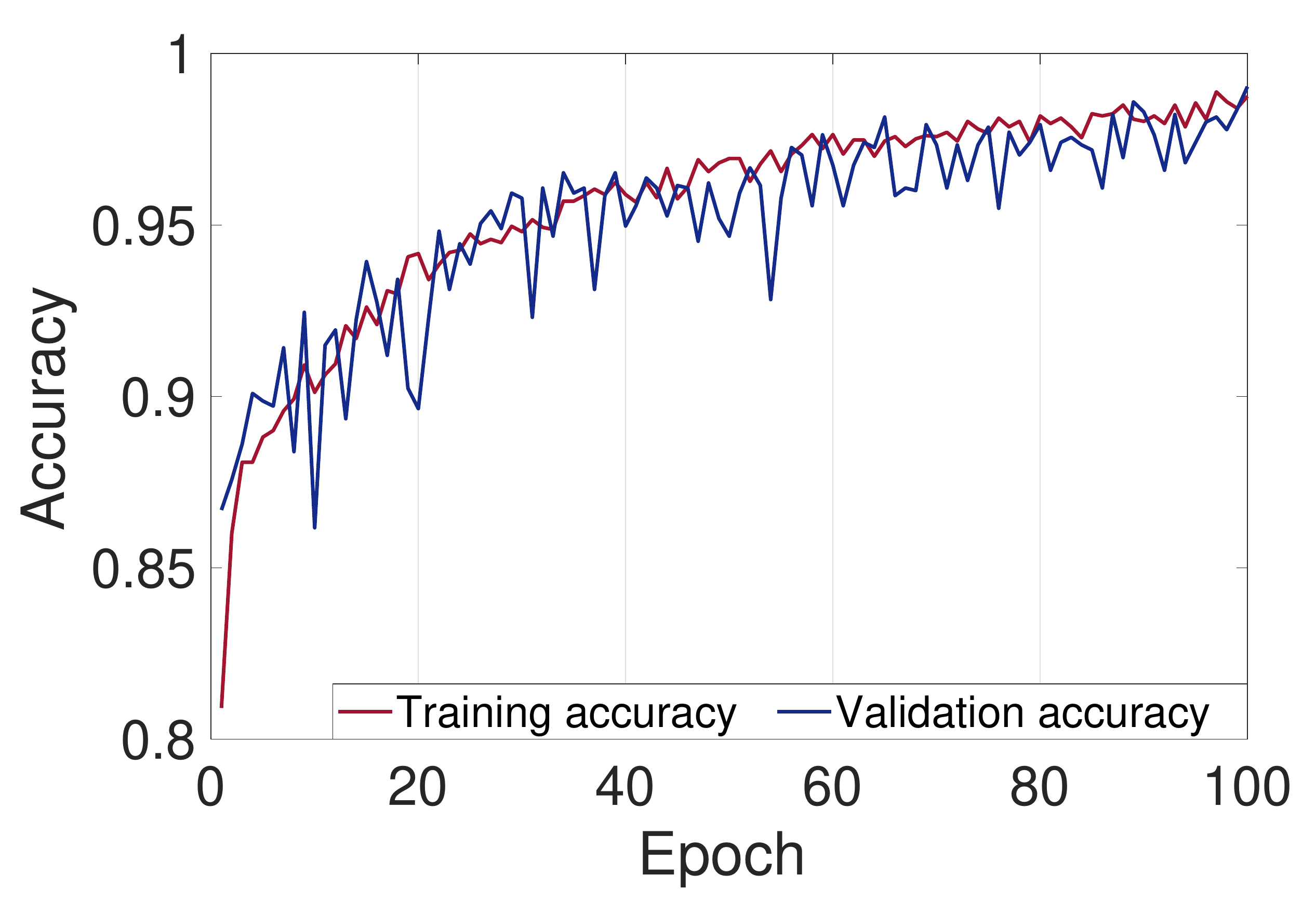}
     \caption{}
     \label{lene_acc}
  \end{subfigure}
  \begin{subfigure}[b]{0.32\linewidth}
    \includegraphics[width=\linewidth]{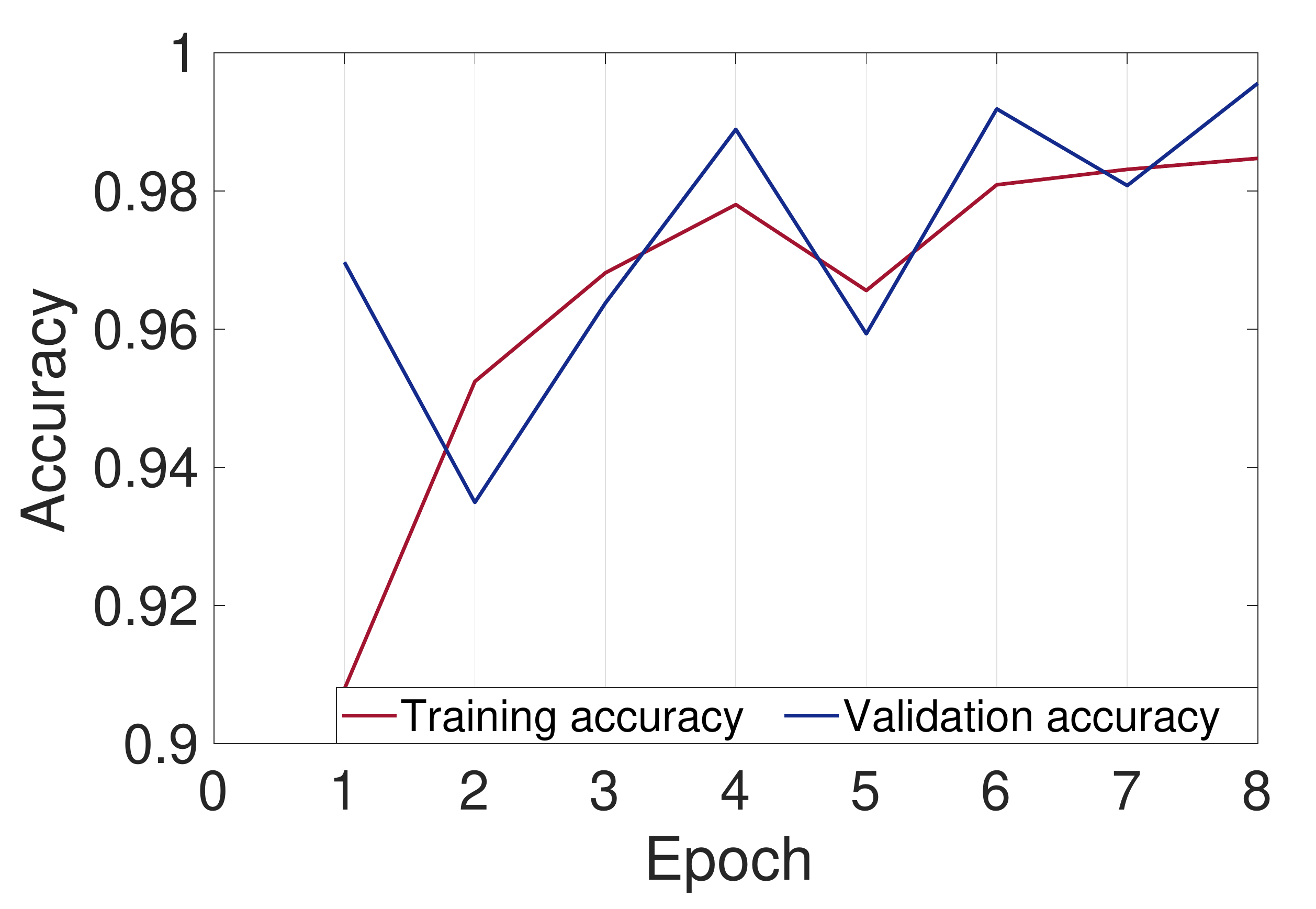}
    \caption{}
    \label{vgg_acc}
  \end{subfigure}
   \begin{subfigure}[b]{0.32\linewidth}
    \includegraphics[width=\linewidth]{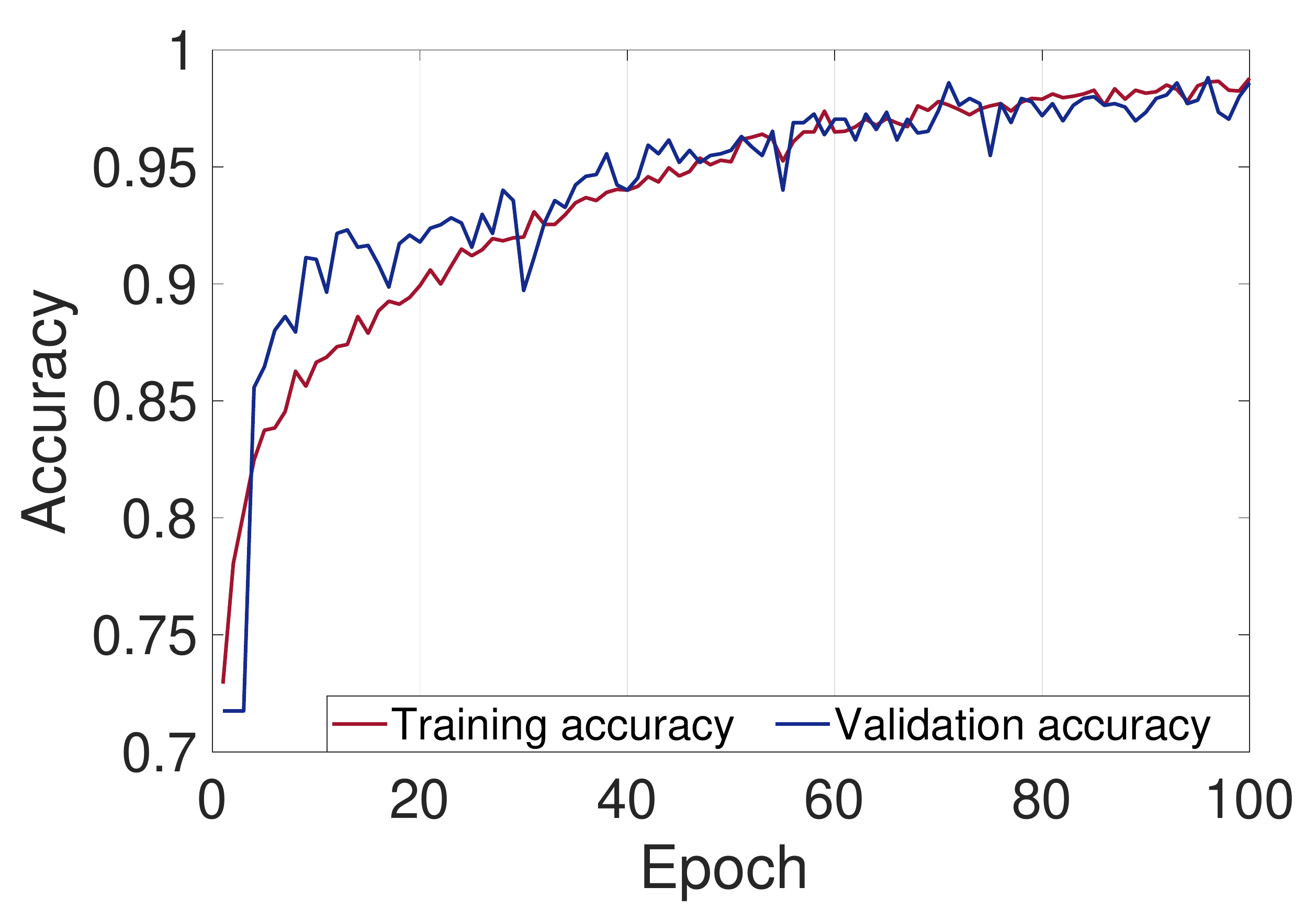}
    \caption{}
    \label{mb2_acc}
  \end{subfigure}
  \caption{Training and test accuracy comparison for (a)~LeNet, (b)~VGG16 and (c)~MobileNetV2 .}
  \label{fig:acc}
\end{figure*}

The training is done over 100 epochs with input image size equal to 52$\times$52 for LeNet-5, VGG16 and MobileNetV2. An early stopping by accuracy approach has been further used to stop the training if the validation accuracy reaches at least the 99.5\%. In this case, only VGG16 has stopped earlier than others (i.e., 8 epochs have been enough) as showed in Figure~\ref{vgg_acc}.

Figure~\ref{fig:acc} shows the accuracy and validation for LeNet, VGG16 and MobileNetV2 over the epochs. Notice that all architectures accuracy increase together with the validation accuracy and converge in almost 0.99. This confirms that the training sessions have been successful without overfitting. Moreover, VGG16 has reached the desired validation accuracy quite earlier than others thanks to its high number of parameters and its deep structure. 
Precisely, LeNet finishes its training with a test precision of 99\%, while VGG16 completes the test precision with 99.5\%, and MobileNet have reached a test precision of 98.5\%. 

\subsection{Validation}
The obtained DNN has been tested over images that have been retrieved from the dataset before the training phase. In this way, it is possible to test the network capability even with new images, thus assess its generalization capability.  As stated before, 900 images have been used for the tests. The results for all three solutions are shown in Table~\ref{tab:res}.

\begin{table}
		\centering
		\caption{LeNet, VGG16 and MobileNetV2 test results.}
		\label{tab:res}
		\resizebox{.9\columnwidth}{!}{%
			\begin{tabular}{c|cccc}
				& \textbf{\begin{tabular}[c]{@{}c@{}}Accuracy\end{tabular}} & \textbf{\begin{tabular}[c]{@{}c@{}}Recall\end{tabular}} & \textbf{\begin{tabular}[c]{@{}c@{}}Precision\end{tabular}} & \textbf{\begin{tabular}[c]{@{}c@{}}F-score\end{tabular}} \\ \hline
				\textbf{LeNet-5} & 96.1                                                                & 94.9                                                               & 99.6                                                                 & 97.2                                                                 \\
				\textbf{VGG16} & 97.9                                                                & 97.4                                                               & 99.6                                                                 & 98.5                                                                  \\
				\textbf{MobileNetV2} & 95.1                                                                & 94.5                                                               & 98.5                                                                 & 96.4                
			\end{tabular}%
		}
		
\end{table}

\begin{figure}[b]
		\centering
		\includegraphics[width=.85\columnwidth]{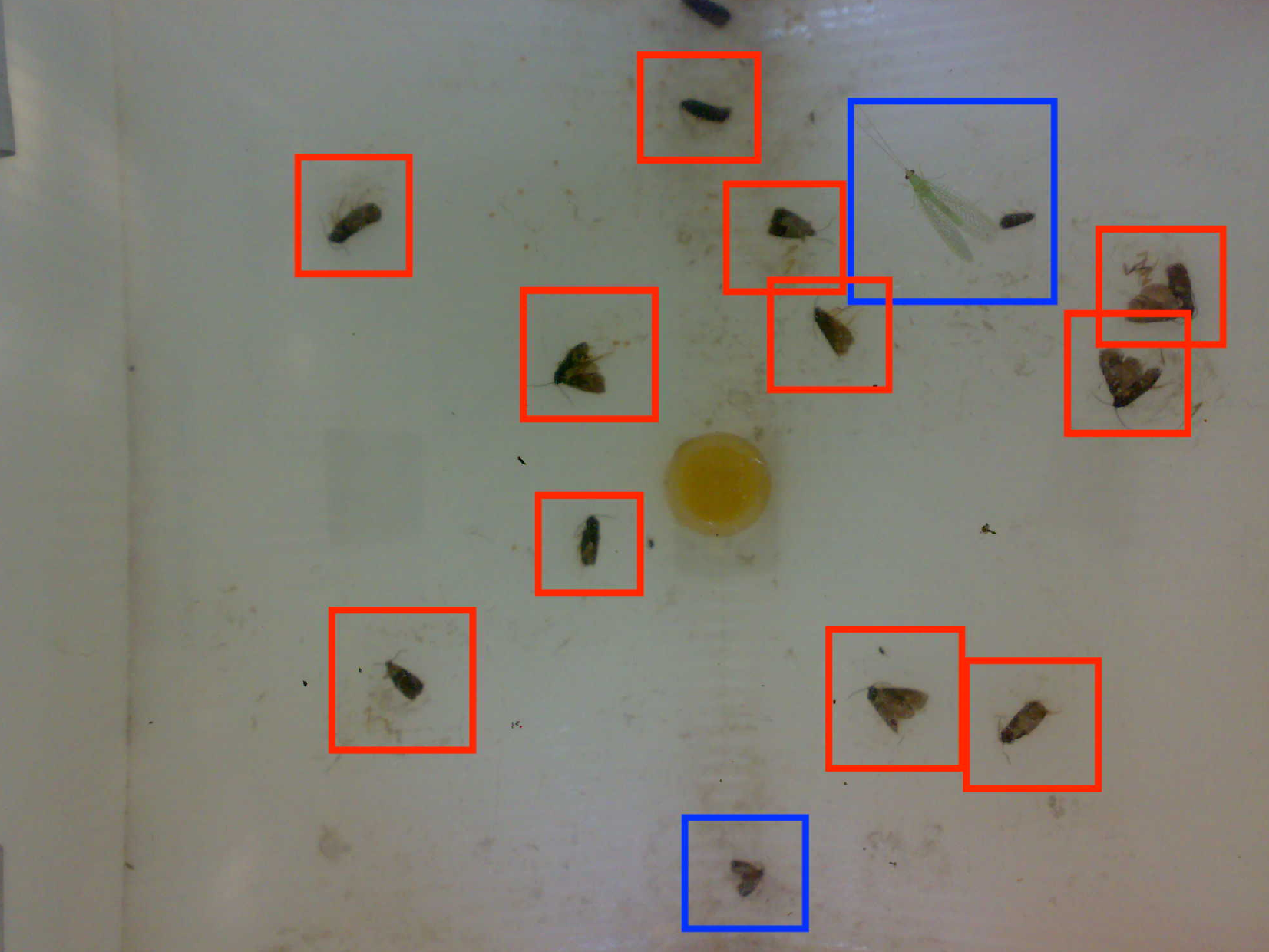}
		\caption{An example of an annotated photo after its evaluation. Red boxes highlights the detected codling moth (positive class) while the blue box general insects (negative class).}
		\label{fig:output}
\end{figure}

Notice that VGG16 features a high precision but a lower recall, which means that it misses some pests, but the predicted ones are accurate, and, consequently, the F-measure is good.
\\
To evaluate if the metrics satisfies the application requirements, we asked both apple farmers and looked into the literature.  Commonly, farmers detect Codling Moth infestation exploiting pheromone-based traps. The traps are then checked once every week, and the pesticide treatment executed when 2 moths/week~\cite{moth1} are trapped/detected. 
Considering the average number of Moth caught seasonally, as reported in several articles~\cite{moth1,moth2}, the accuracy of the system is perfectly suitable for automating the Codling Moth detection task. Moreover, the smart trap is programmed to check for Codling Moth twice per day, allowing a prompter reaction compared to today's human-inspection approach.  
\\
All architectures confirm their generalization capability if new images are given as input. Thus, it implies that the models have been trained without overfitting. On the other hand, both LeNet and MobileNetV2 present good results in all parameters. Thus, it is possible to state that all three architectures can be used in this application scenario. However, considering the deep architecture and its high number of parameters, VGG16 and MobileNet could outperform LeNet if the training dataset is even more consistent. For reference, an example of moth detection in a real scenario is showed in Figure~\ref{fig:output} where codling moth are highlighted with red bounding boxes.

\subsection{Network Optimization and Data augmentation}
    Much of the success of deep learning has come from building larger and larger neural networks. This allows these models to perform better on various tasks, but also makes them more expensive to use. Larger models take more storage space which makes them harder to distribute. Larger models also take more time to run and can require more expensive hardware. The optimization phase aims to reduce the size of models while minimizing loss in accuracy or performance. This allows the speedup of the evaluation with minimizing the accuracy loss. The proposed implementation applies optimization both during the training phase and before the evaluation of the training model. 
    \\
    \textbf{Data Augmentation} Training machine learning models, usually requires larger dataset to achieve better performance generalization. In our case, the amount of training data, which is represented by the number of training patches, is much smaller than standard small-scale image classification datasets~\cite{lecun1998mnist, krizhevsky2009learning}, which have on the order of 50,000 training examples. Therefore, we performed data augmentation to increase the number of images for training and incorporate invariance to basic geometric transformations into the classifier. Based on the “top-view” nature of the trap images, a certain patch will not change its class label when it is slightly translated, flipped, or rotated. Therefore, we apply these simple geometric transformations to the original patches to increase the number of training examples.
    \\
    \textbf{Pruning.} Neural network pruning is a method of compression that involves removing unnecessary neurons or weights from a trained model. There are different ways to prune a neural network. 1) You can prune weights, by setting individual parameters to zero and making the network sparse; 2) You can remove entire nodes from the network, making the network architecture itself smaller, while aiming to keep the accuracy of the initial larger network. To not impact the accuracy performance too much, in our case, we prune weights. Network pruning is executed iteratively during the training phase, to achieve the desired accuracy during the validation phase.
    \\
    \textbf{Model Optimization.} After the training phase, the model is further optimized to reduce the network complexity and to increment the evaluation speed. To facilitate faster inference of the deep learning models, we do Node Merging \cite{Fisher1987}, Constant and Horizontal Fusion \cite{li2020automatic}, Batch Normalization \cite{pmlr-v37-ioffe15}, and we drop unused layers (Dropout layer used during the training phase).

\section{Results and Evaluation}
\label{res}
For this application purpose, it is enough to check the presence of codling moth twice per day. In one application cycle, the smart trap has:
\begin{itemize}
    \item 
        \textbf{[Boot]} - Power ON
    \item
        \textbf{[Task 1]} - Take a picture of the trapped insects
    \item
        \textbf{[Task 2]} - Pre-process the capture images
    \item
        \textbf{[Task 3]} - Execute the classification algorithm
    \item
        \textbf{[Task 4]} - Send the computation results using the radio
    \item
        \textbf{[Shutdown]} - Power OFF
\end{itemize}
These steps have been used to characterize the smart trap performance both in terms of power consumption and requiring energy. Moreover, since the system has preferably to work unattended indefinitely, few remarks and considerations about its energy balance have been taken into account.

\begin{table*}[h]
\centering
\caption{Energy tasks breakdown of the tested platforms. The best trade-off is provided by Raspberry Pi3 evaluating LeNet network.}
\label{tab:energy}
\resizebox{\textwidth}{!}{%
\begin{tabular}{c|cccccccccccc}
\textbf{}         & \multicolumn{12}{c}{\textbf{Platform}}                                                                                                                                                                                                                                                                                                                                                                                                                                                    \\ \hline
\textbf{Task}     & \textbf{\begin{tabular}[c]{@{}c@{}}RPi3 \\ MobileNetv2\end{tabular}} & {\color[HTML]{009901}\textbf{\begin{tabular}[c]{@{}c@{}}RPi3 \\ LeNet\end{tabular}}} & \textbf{\begin{tabular}[c]{@{}c@{}}RPi3 \\ VGG16\end{tabular}} & \textbf{\begin{tabular}[c]{@{}c@{}}RPi3 MobileNet \\ with NCS\end{tabular}} & \textbf{\begin{tabular}[c]{@{}c@{}}RPi3 LeNet \\ with NCS\end{tabular}} &  \textbf{\begin{tabular}[c]{@{}c@{}}RPi3 VGG16 \\ with NCS\end{tabular}} & {\color[HTML]{CB0000} \textbf{\begin{tabular}[c]{@{}c@{}}RPi4 \\ MobileNetV2\end{tabular}}} & \textbf{\begin{tabular}[c]{@{}c@{}}RPi4 \\ LeNet\end{tabular}} & \textbf{\begin{tabular}[c]{@{}c@{}}RPi4 \\ VGG16\end{tabular}} & \textbf{\begin{tabular}[c]{@{}c@{}}RPi4 MobileNetV2 \\ with NCS\end{tabular}} & \textbf{\begin{tabular}[c]{@{}c@{}}RPi4 LeNet \\ with NCS\end{tabular}} & \textbf{\begin{tabular}[c]{@{}c@{}}RPi4 VGG16 \\ with NCS\end{tabular}}\\ \hline
\textbf{Boot [J]}     & 40.7                                                                 & {\color[HTML]{009901} 40.7}                                                                  & 40.7                                                          & 40.7                                                          & 40.7                                                                   & 40.7     & {\color[HTML]{CB0000} 56}        & 56       & 56      & 56
& 56    & 56    \\
\textbf{Task 1 [J]}   & 2.171                                                                   & {\color[HTML]{009901} 1.674}                                                                   & 1.733                                                           & 2.427                                                           & 2.177                                                                    & 2.359                               & {\color[HTML]{CB0000} 1.327}   & 2.179     & 1.907     & 3.385     & 1.934     & 2.609                            \\
\textbf{Task 2 [J]}   & 6.125                                                                   & {\color[HTML]{009901} 5.453}                                                                   & 6.491                                                           & 6.923                                                           & 7.054                                                                    & 7.041                               & {\color[HTML]{CB0000} 5.099}   & 5.139     & 5.221     & 6.257     & 6.37      & 6.423                           \\
\textbf{Task 3 [J]}   & 119.1                                                                 & {\color[HTML]{009901} 57.4}                                                                  & 114.3                                                          & 70.34                                                          & 59.04                                                                   & 73.53                                    & {\color[HTML]{CB0000} 111}   & 49.39     & 75.22     & 90.2      & 68.28     & 66.33                      \\
\textbf{Task 4 [J]}   & 2.147                                                                   & {\color[HTML]{009901} 2.147}                                                                   & 2.147                                                           & 2.273                                                           & 2.273                                                                    & 2.273                             & {\color[HTML]{CB0000} 1.822}     & 1.822     & 1.822     & 1.822     & 1.822     & 1.822                         \\
\textbf{Shutdown [J]} & 15.85                                                                  & {\color[HTML]{009901} 15.85}                                                                  & 15.85                                                          & 21.97                                                          & 21.97                                                                   & 21.97      
& {\color[HTML]{CB0000} 24.84}     & 24.84     & 24.84     & 24.84     & 24.84     & 24.84      \\ \hline
\textbf{Total [J]}    & \textbf{186.1}                                                        & {\color[HTML]{009901} \textbf{123.2}}                                                        & \textbf{181.2}                                                & \textbf{144.6}                                                & \textbf{133.2}                                                         & \textbf{147.9}      & {\color[HTML]{CB0000} \textbf{200.1}}    & \textbf{139.4}     & \textbf{165}   & \textbf{182.5}     & \textbf{159.2}     & \textbf{158}                                                  
\end{tabular}%
}
\end{table*}

\subsection{Power consumption}
\begin{figure}[b]
	\centering
	\includegraphics[width=1\linewidth]{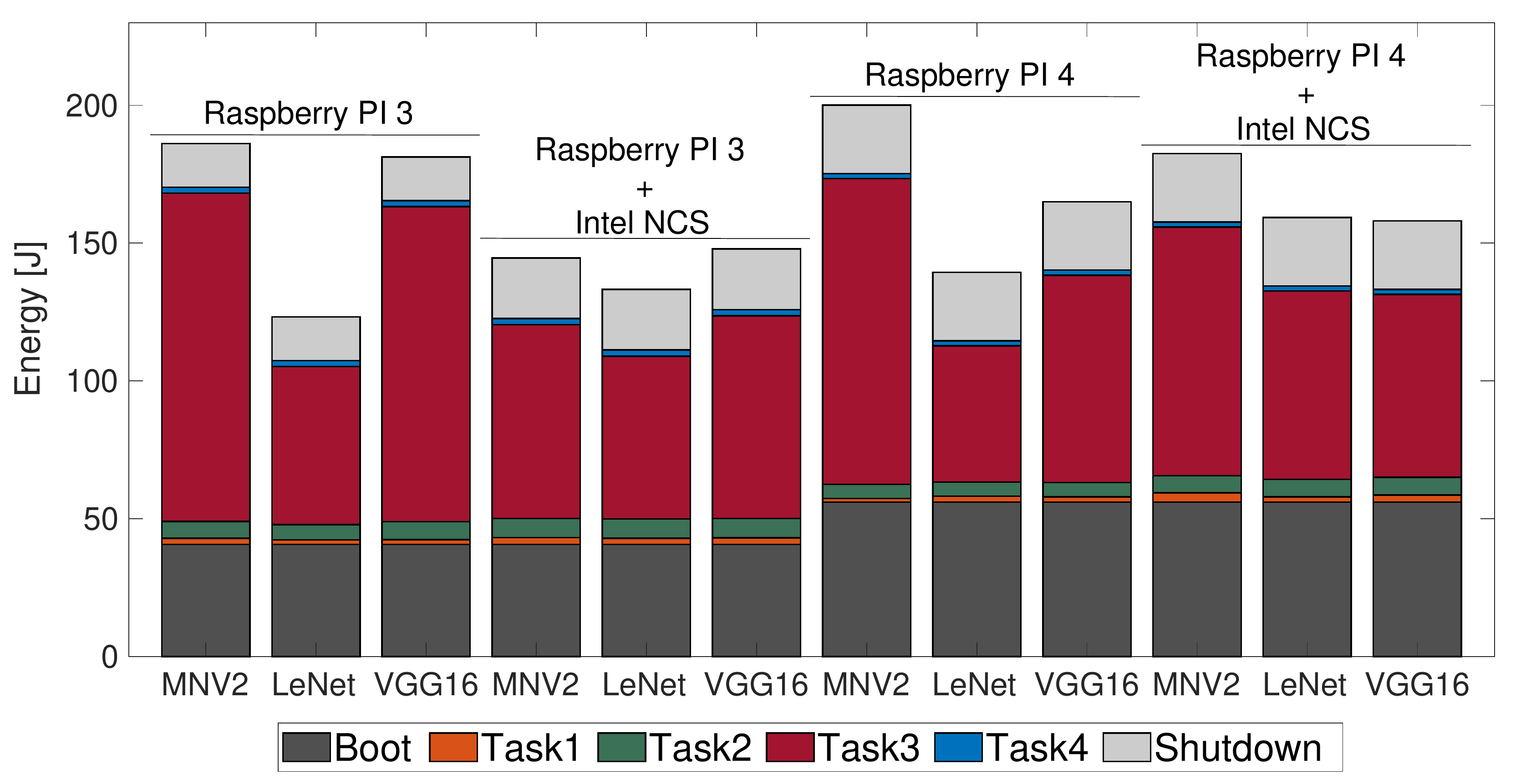}
	\caption{Tasks Energy consumption breakdown comparison for a single cycle of the implemented application. Single-task energy is presented in table~\ref{tab:energy} for each implementation.}
	\label{fig:perf_car}
\end{figure}

Thanks to the external low-power microcontroller, the system can be waked up only when planned to perform its tasks. To characterize the system consumption, the current required by each task has been considered by comparing through twelve different configurations:

\begin{itemize}
    \item Raspberry Pi3 evaluating MobileNet V2;
    \item Raspberry Pi3 evaluating LeNet;
    \item Raspberry Pi3 evaluating VGG16;
    \item Raspberry Pi3 supported by NCS evaluating MobileNet;
    \item Raspberry Pi3 supported by NCS evaluating LeNet;
    \item Raspberry Pi3 supported by NCS evaluating VGG16;
    \item Raspberry Pi4 evaluating MobileNet V2;
    \item Raspberry Pi4 evaluating LeNet;
    \item Raspberry Pi4 evaluating VGG16;
    \item Raspberry Pi4 supported by NCS evaluating MobileNet;
    \item Raspberry Pi4 supported by NCS evaluating LeNet;
    \item Raspberry Pi4 supported by NCS evaluating VGG16;
   
\end{itemize}
These configurations permit finding the software bottlenecks and selecting the best trade-off for the overall system performance.
As can be noted in Table~\ref{tab:energy}, Task3 is the one that requires more energy because of the neural inference computation together with the use of NCS. Even if it boosts the neural inference step, its VPU requires a high power to perform accelerated edge inference. Here, the best arrangement that absorbs less power on average than others is the Raspberry Pi3 evaluating LeNet-5. Figure~\ref{fig:perf_car} shows the energy consumption of each task. In this case, Task3 is confirmed as the most power-hungry activity and requires more energy than others. Moreover, configurations that involve Raspberry Pi4 are characterized by higher energy consumption than others. This is due to the Pi4, which features oversized CPU performance. Even if it should execute the application faster than Pi3, it requires more energy to complete one application cycle. Thus, the configuration consisting of the Raspberry Pi3 running LeNet demonstrates to provide the best trade-off.

\subsection{Expected Battery Lifetime}
The platform is powered by a single cell LiPo battery with 1820 mAh capacity. Considering the most energy-demanding configuration (i.e., Raspberry Pi4 evaluating MobileNetV2 consuming 200.1 J for one application cycle), the battery could supply the system for about 120 complete cycles, which last around two months. On the other hand, if the best configuration is considered (i.e., Raspberry Pi3 evaluating LeNet consuming 123.2 J for one application cycle), the system operates unattended for about 200 cycles (around 3 months) with the only battery. 

\begin{figure*}
  \centering
  \begin{subfigure}[b]{0.32\linewidth}
    \includegraphics[width=\linewidth]{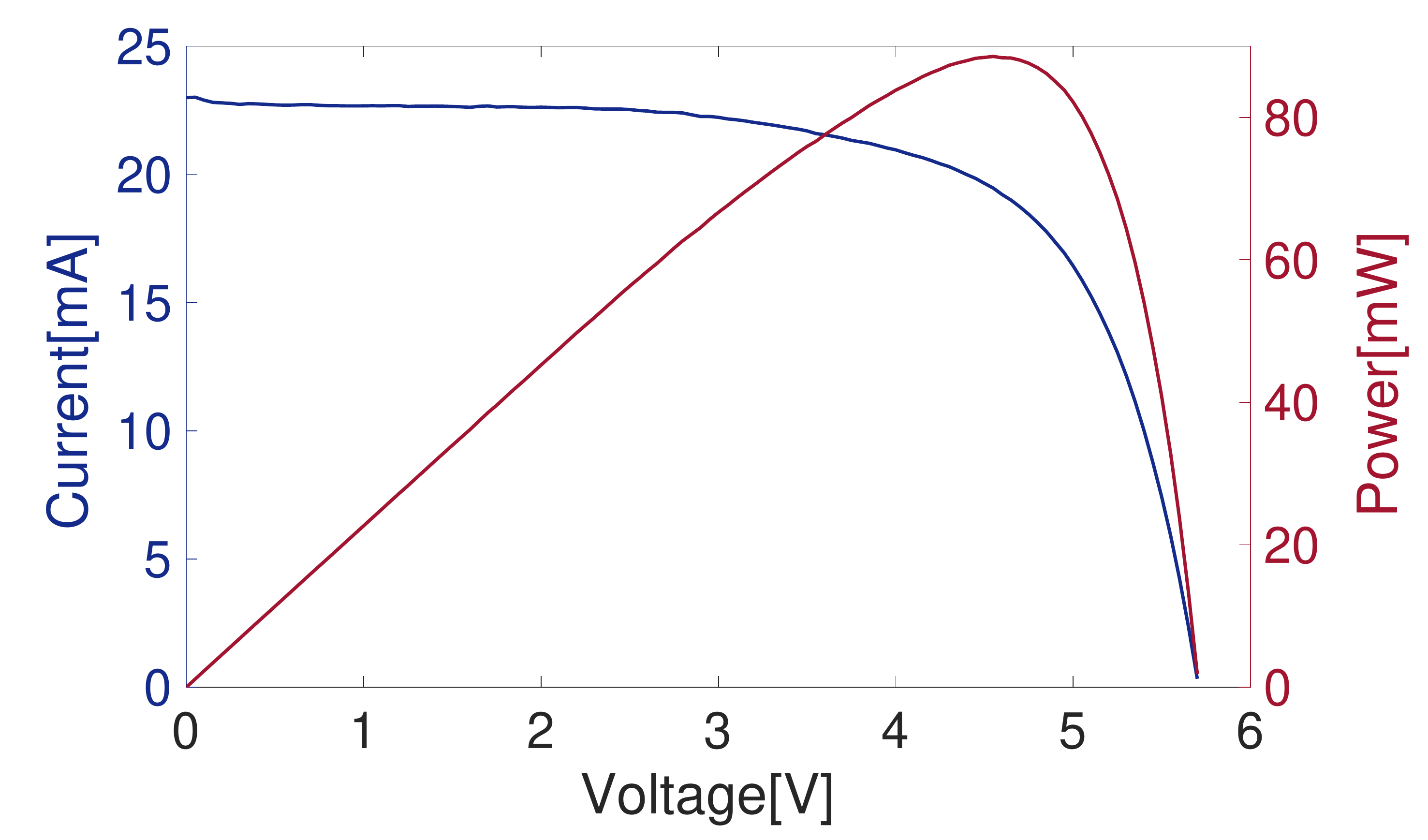}
     \caption{}
     \label{2000}
  \end{subfigure}
  \begin{subfigure}[b]{0.32\linewidth}
    \includegraphics[width=\linewidth]{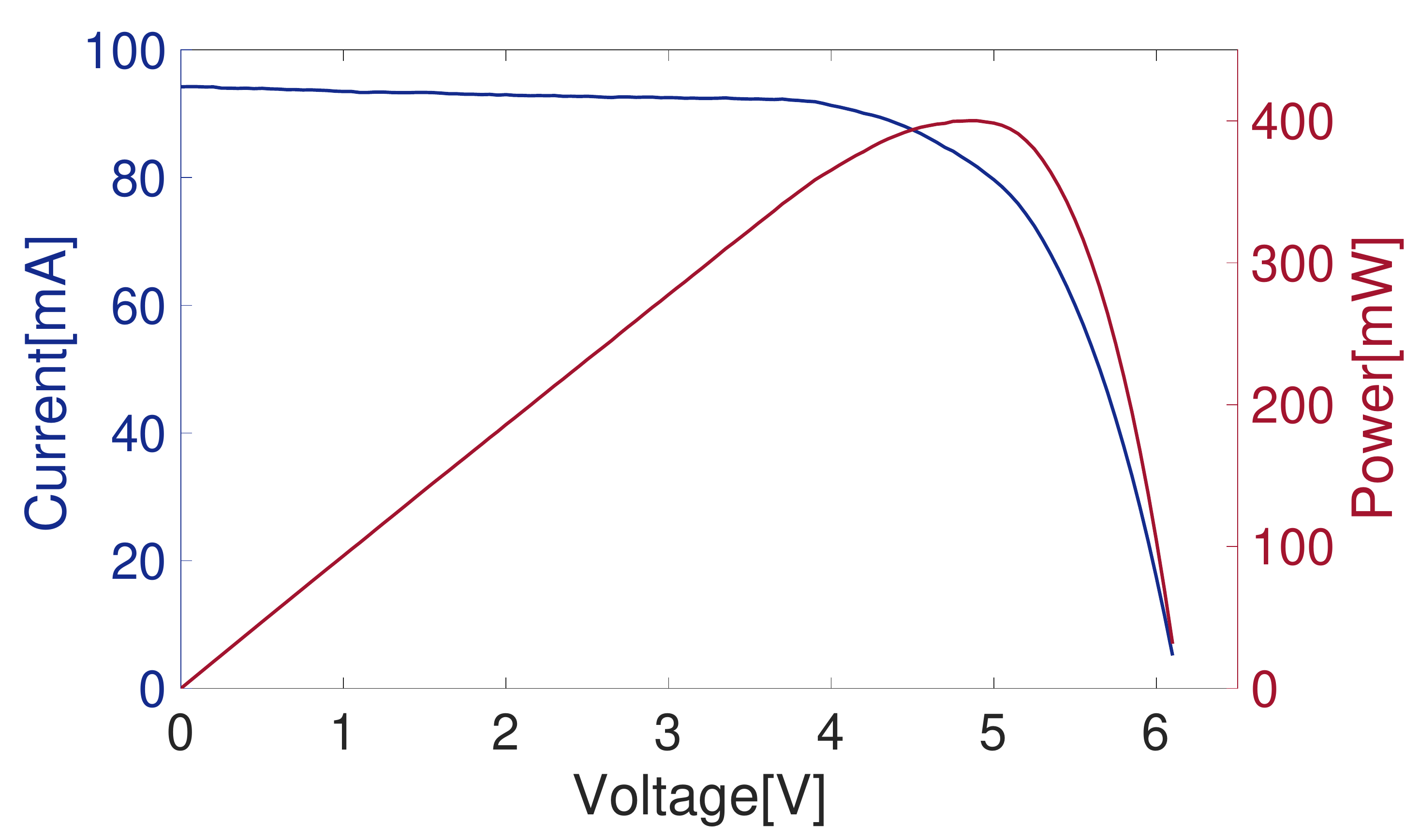}
    \caption{}
    \label{10000}
  \end{subfigure}
  \begin{subfigure}[b]{0.32\linewidth}
    \includegraphics[width=\linewidth]{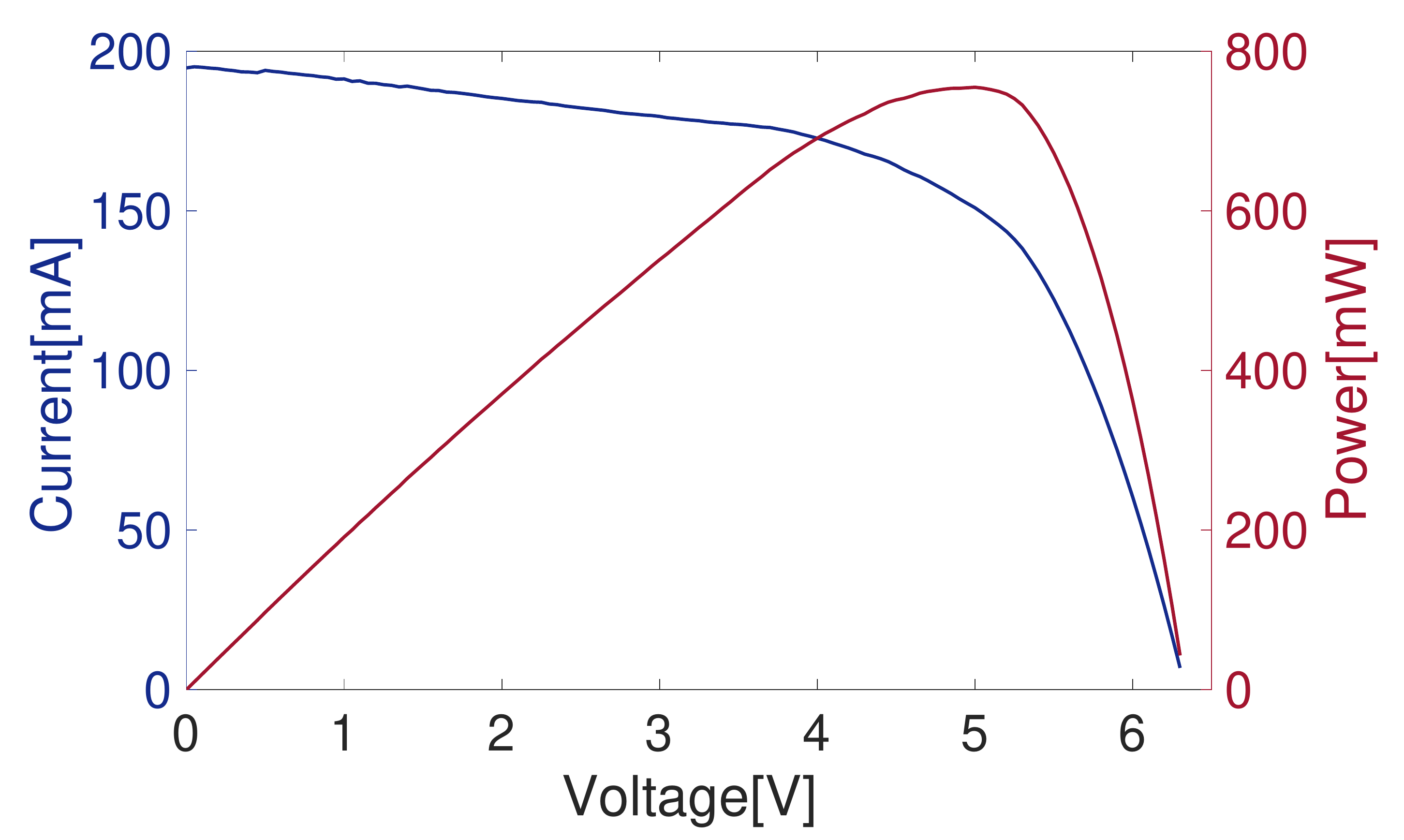}
    \caption{}
    \label{25000}
  \end{subfigure}
  \caption{Solar panel characterization respectively (a) at 2000 lx, (b) at 10 klx and (c) at 25 klx.}
  \label{fig:solar}
\end{figure*}

\subsection{Energy Harvesting and Platform Sustainability}
\begin{figure}
		\centering
		\includegraphics[width=1\columnwidth]{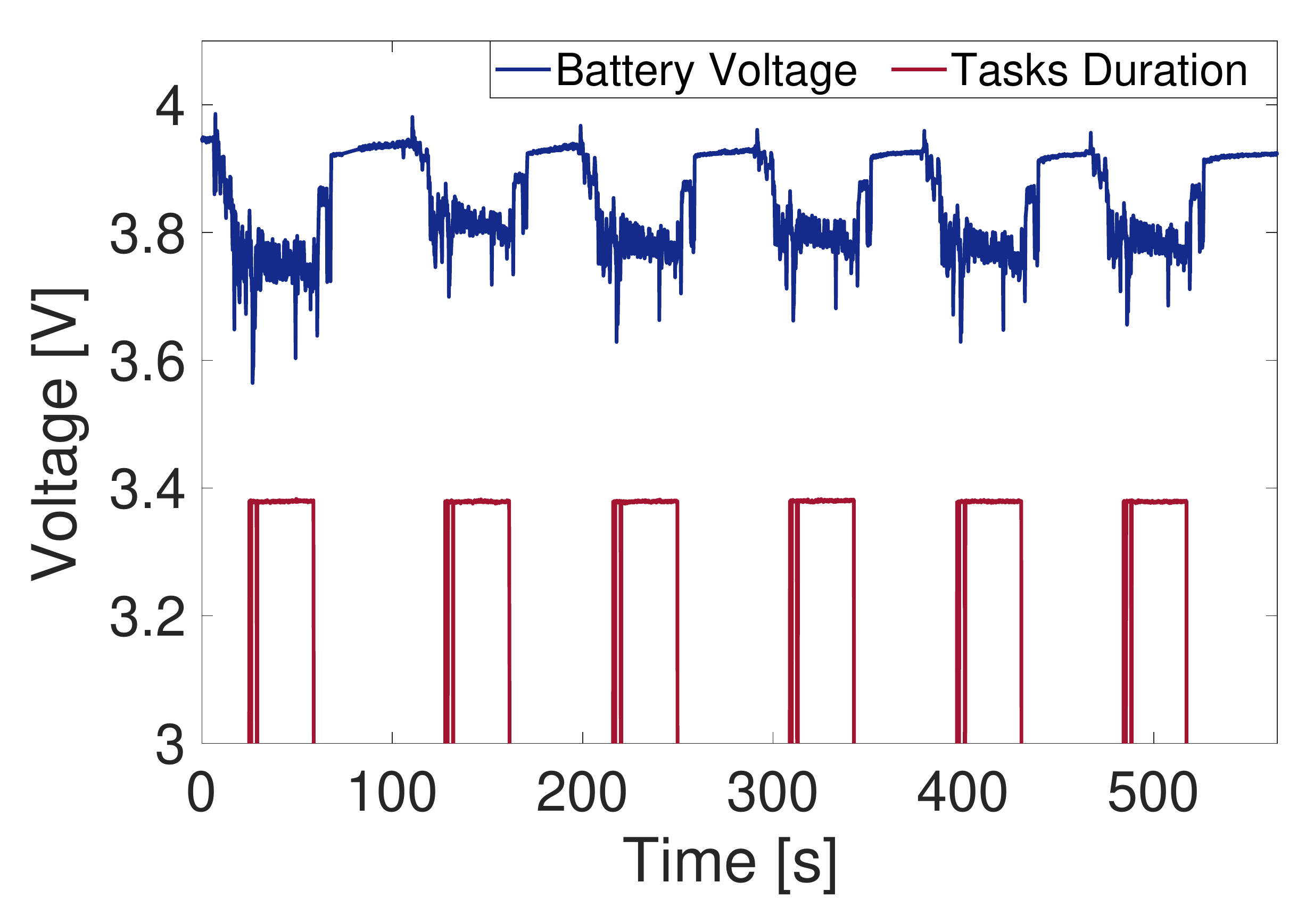}
		\caption{Platform sustainability evaluation. The graphs presents the battery voltage trend while harvesting solar energy with a 7000 lx illuminance.}
		\label{fig:energy_harv}
\end{figure}

In apple orchards, energy resources are limited. Thus a   self-sustainable system leads to a more robust and durable application to foster farmers to "deploy and forget" the sensors. For this purpose, an energy harvesting system has been designed and characterized. It is composed of a MCU to trigger the power-on and shut down and a 140mm$\times$100mm solar panel used for recharging the battery. The solar panel used has been characterized using several different light levels. Figure~\ref{fig:solar} present the I-V curves of the solar panel in three different environmental situations: 2000 lx (i.e., cloudy, Figure~\ref{2000}), 10000 lx (i.e., fair, Figure~\ref{10000}) and 25000 lx (i.e., sunny, Figure~\ref{25000}).
\\
By considering the measured power output from the solar harvester, we measured the time duration of a whole charge process of the battery in different conditions. We have measured both the time needed for charging a depleted battery, and the time to harvest the energy necessary for a single application cycle. The results are presented in Table~\ref{tab:solar}. As can be noted, already starting with the lower illuminance level, the recharging time is lower than the application duty cycle (i.e., 2 cycle per day), validating the sustainability of the platform. This feature is also confirmed by the graphs presented in Figure~\ref{fig:energy_harv} that show the battery energy trend while harvesting with an illuminance equal to 7000~lx. It can be argued that even by emulating 3 days of cycles, the battery level does not drop, validating the hypothesis that the platform can self sustain its operation thanks to the integrated solar harvester.

\begin{table}
		\centering
		\caption{Battery recharge time for both a single application cycle and for fully charge the battery [20-100\%] while harvesting solar energy at three different illuminance level.}
		\label{tab:solar}
		\resizebox{.9\columnwidth}{!}{%
			\begin{tabular}{c|ccc}
				& \textbf{\begin{tabular}[c]{@{}c@{}}2000 lx\end{tabular}} & \textbf{\begin{tabular}[c]{@{}c@{}}10000 lx\end{tabular}} & \textbf{\begin{tabular}[c]{@{}c@{}}25000 lx\end{tabular}} \\ \hline
				\textbf{Full Battery} & 60 h                                                                & 13 h                                                               & 7 h                                                                  \\
				\textbf{Single Cycle} & 23 min                                                                & 5 min                                                               & 3 min     
			\end{tabular}%
		}
\end{table}

\section{Conclusions}
\label{close}
Computer vision systems are already widely employed in different segments of precision agricultural and industrial food production. Running deep learning features on the edge can optimize the management of fruit orchards.

This paper presents a computer vision solution for automating pest detection inside orchards. The platform exploits ML functionalities on edge to evaluate images captured inside common pheromone traps to get early detection of dangerous parasites. Furthermore, on board inference avoids the transmission of the whole images, reducing the wireless communication bandwidth and energy costs.
We analyzed the best hardware configuration using different neural networks, trained to get the best pest detection accuracy. Moreover, we combined a designed energy harvester to demonstrates the perpetual operation of the device unattended. 




\begin{IEEEbiography}[{\includegraphics[width=1in,height=1.25in,clip,keepaspectratio]{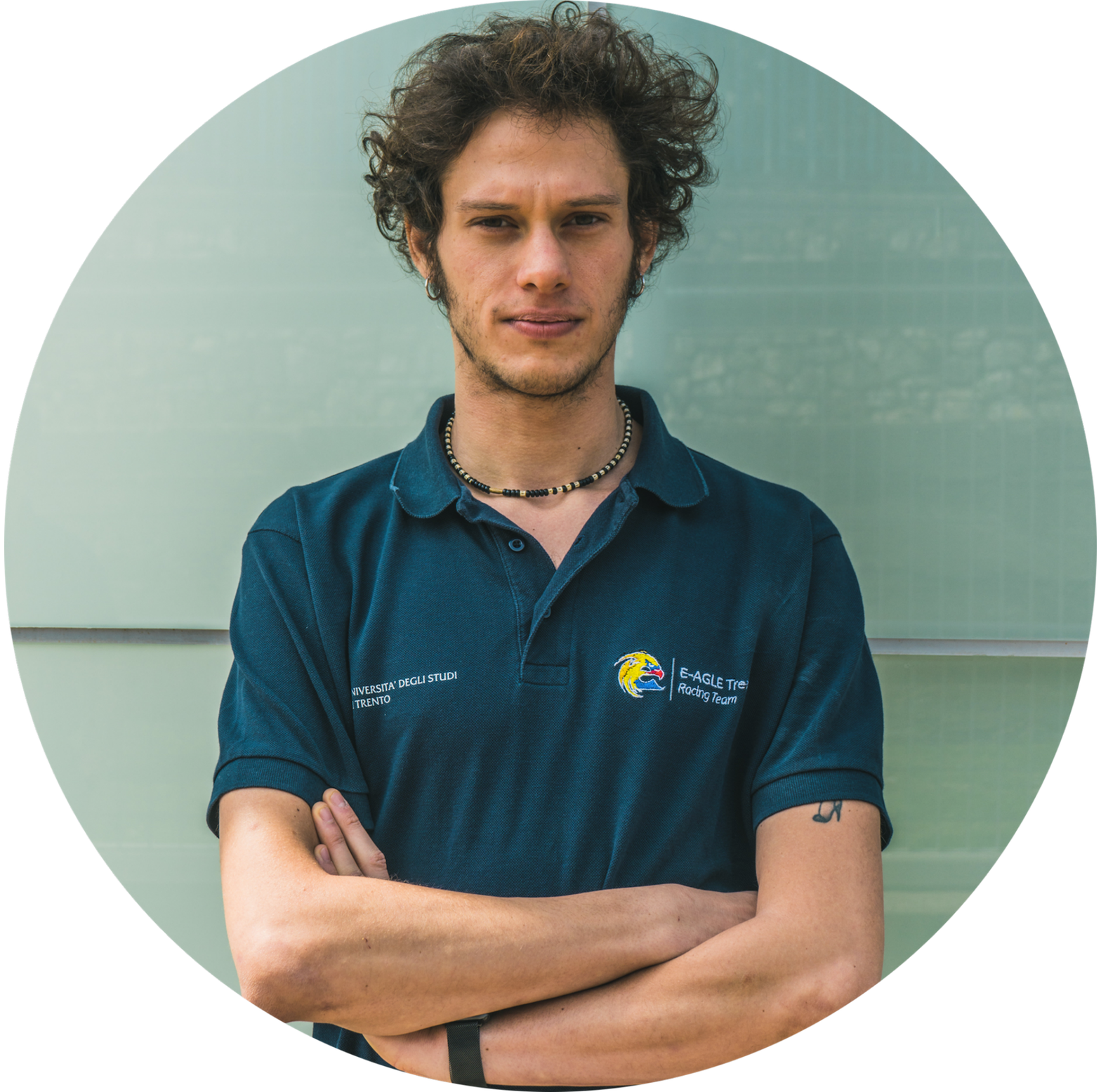}}]{Andrea Albanese}
received his B.S. in electrical and telecommunication engineering and his M.S. in mechatronics engineering with electronics and robotics specialization from University of Trento in 2017 and 2020 respectively. He is currently a fellow researcher at the same university with focus in machine learning optimization techniques for resource constrained environment (e.g., MCUs) applied on UAVs. He was involved for three years in the university Formula Student team with duties on power systems and embedded systems. His major areas of interest are tiny machine learning and optimization techniques, sensor fusion and autonomous navigation. 
\end{IEEEbiography}

\vskip 0pt plus -1fil

\begin{IEEEbiography}[{\includegraphics[width=1in,height=1.25in,clip,keepaspectratio]{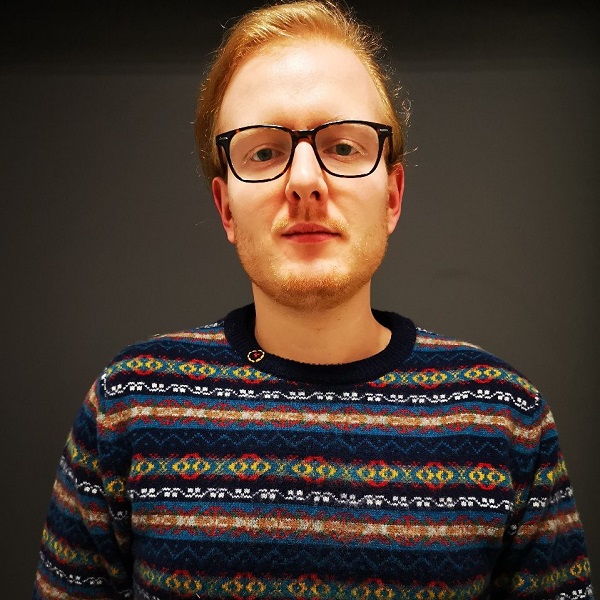}}]{Matteo Nardello}
 is currently a Postdoc researcher at the Department of Industrial Engineering, University of Trento. He obtained his PhD in Systems Engineering in 2020 and M.S in the field of Electronics and Telecommunications Engineering in 2016. His research interests include the investigation of machine learning techniques applied to resource constrained embedded platforms, with a special focus on autonomous smart IoT devices and the study of new architecture for indoor localization services. His other research interests encompass modeling and hardware-software co-design of innovative solutions to reduce power requirements of distributed wireless sensors network for environmental sensing. 
\end{IEEEbiography}

\vskip 0pt plus -1fil

\begin{IEEEbiography}[{\includegraphics[width=1in,height=1.25in,clip,keepaspectratio]{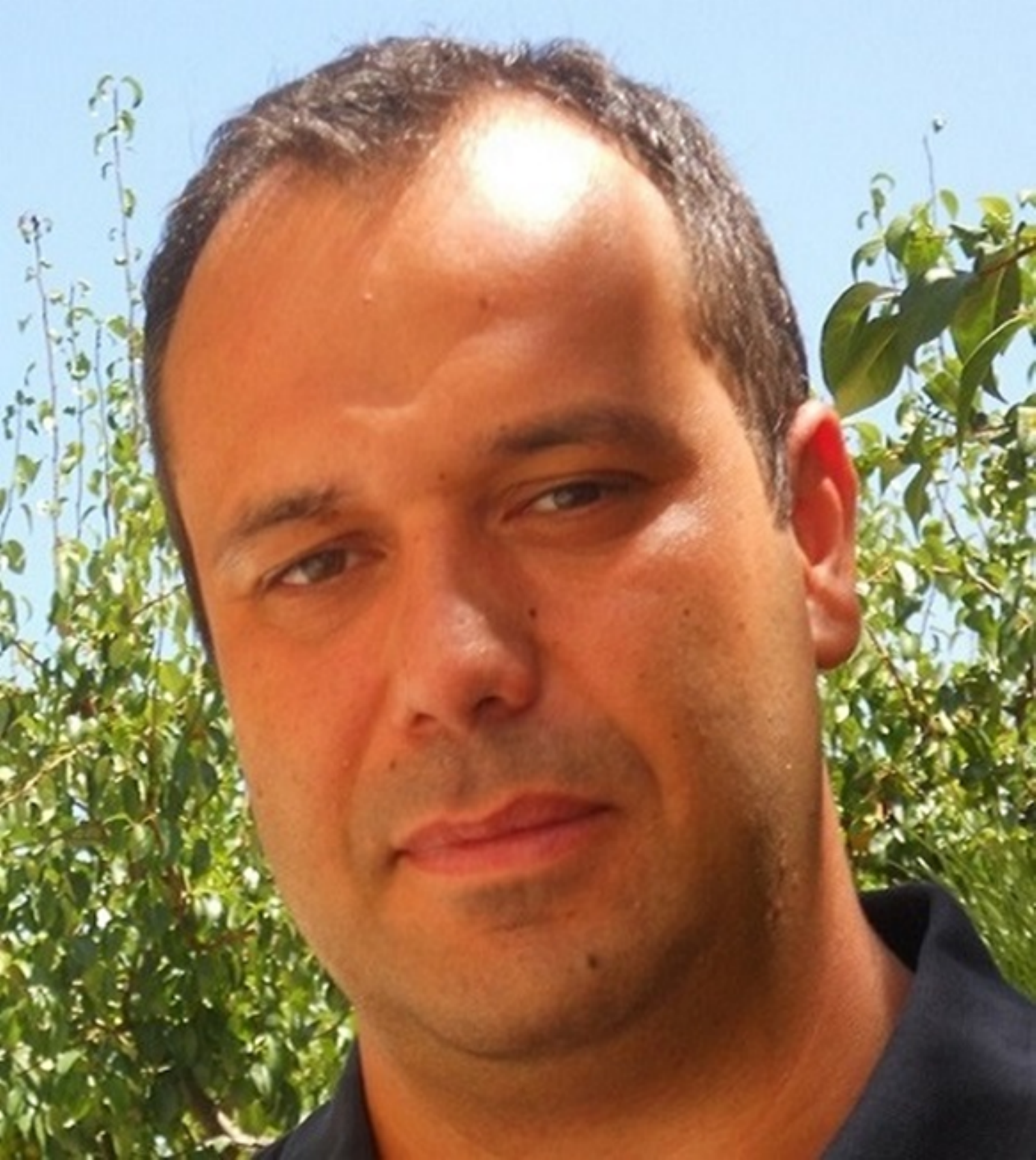}}]{Davide Brunelli} (Senior Member, IEEE) received the M.S. (cum laude) and Ph.D. degrees in electrical engineering from the University of Bologna, Bologna, Italy, in 2002 and 2007, respectively. He is currently an associate professor at the University of Trento, Italy. His research interests include IoT and distributed lightweight unmanned aerial vehicles, the development of new techniques of energy scavenging for low-power embedded systems and energy-neutral wearable devices, Drones, UAVs and Machine Learning. He was leading industrial cooperation activities with TIM Italy, ENI, and STMicroelectronics. He has published more than 200 papers in international journals or proceedings of international conferences. He is an ACM member.
\end{IEEEbiography}

\balance

\end{document}